%% file: main.tex
\documentclass{article}

\PassOptionsToPackage{numbers, compress}{natbib}
\usepackage[main, final]{neurips_2026}

\input{preamble}
\usepackage[utf8]{inputenc}
\usepackage[T1]{fontenc}
\usepackage{hyperref}
\usepackage{url}
\usepackage{booktabs}
\usepackage{amsfonts}
\usepackage{nicefrac}
\usepackage{microtype}
\usepackage[table,xcdraw,dvipsnames]{xcolor}
\usepackage{multirow}
\usepackage{amsmath}
\usepackage{graphicx}
\usepackage{caption}
\usepackage{algorithm}
\usepackage{algorithmic}
\usepackage{amssymb}   
\usepackage{wrapfig}   
\usepackage{placeins}  
\usepackage{float}     

\title{SRA: Spatial Reasoning Adapter via Evolving Social Interaction Graphs for Trajectory Prediction}

\author{%
  Jaewoo Jeong$^{1}$\thanks{Equal contribution.} \quad
  Seonkyu Song$^{1}$\footnotemark[1] \quad
  Hyeonwoo Park$^{2}$ \quad
  Jegyeong Cho$^{1}$ \\
  \textbf{Yujin Bae$^{1}$} \quad
  \textbf{Giwon Lee$^{1}$} \quad
  \textbf{Daehee Park$^{2,3}$} \quad
  \textbf{Kuk-Jin Yoon$^{1}$} \\
  $^{1}$KAIST \quad $^{2}$DGIST \quad $^{3}$UNIVA \\
  \texttt{\{jeong207, sok9854, j2k0618, yujinbae, dlrldnjs, kjyoon\}@kaist.ac.kr} \\
  \texttt{\{hyeonu4995, dhpark\}@dgist.ac.kr}
}

\hypersetup{pdftitle={SRA: Spatial Reasoning Adapter via Evolving Social Interaction Graphs for Trajectory Prediction},
  pdfauthor={Jaewoo Jeong, Seonkyu Song, Hyeonwoo Park, Jegyeong Cho, Yujin Bae, Giwon Lee, Daehee Park, Kuk-Jin Yoon}}

\begin{document}

\maketitle

\begin{center}
    \centering
    \captionsetup{type=figure}
    \includegraphics[width=.98\textwidth]{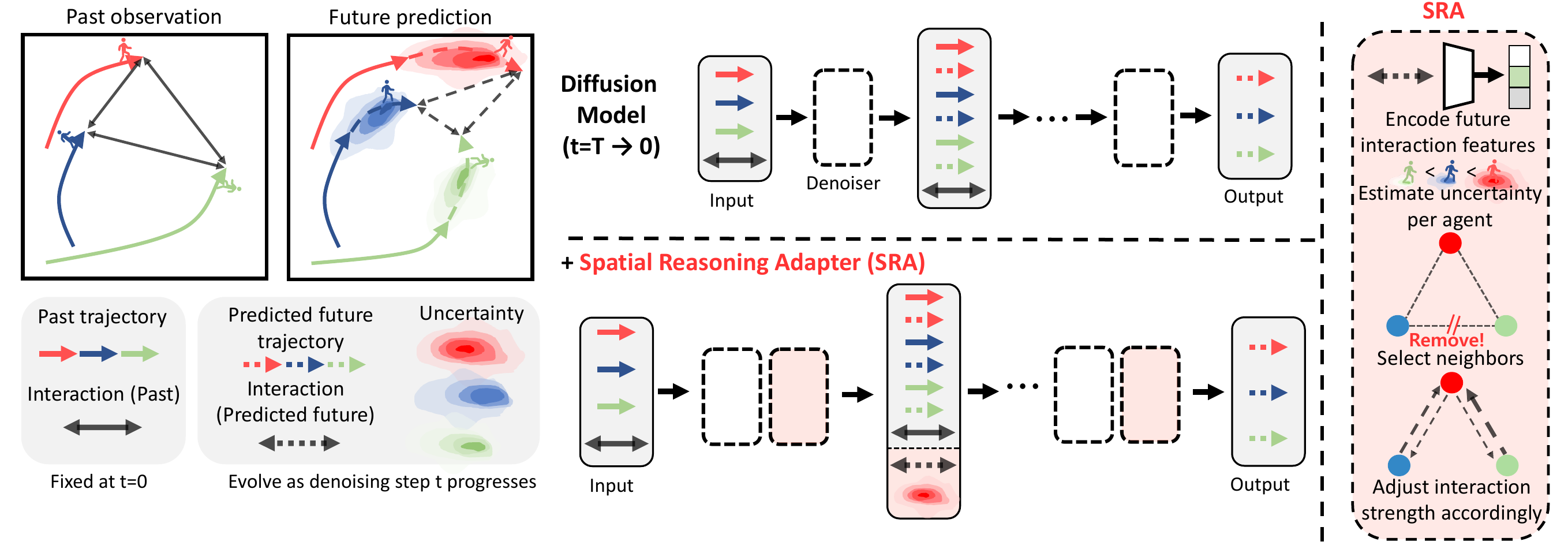}
    \vspace{-8pt}
    \caption{\textbf{SRA} (\textbf{S}patial \textbf{R}easoning \textbf{A}dapter). \emph{Left:} conventional predictors encode interactions (black arrows) once from the past trajectories and keep them fixed during denoising, whereas SRA infers interactions among the \emph{predicted} futures, together with a per-agent uncertainty, and lets them evolve as denoising progresses. \emph{Middle:} a host diffusion model (top) and the same model with SRA inserted at every denoising step (bottom, red blocks). \emph{Right:} at each step, SRA encodes future interaction features, estimates per-agent uncertainty, selects neighbors (removing irrelevant edges), and adjusts interaction strength by uncertainty, refining spatial relations for coherent scene-level forecasts.}
    \label{fig:teaser}
\end{center}
\vspace{-6pt}

\input{sec/0_abstract}    
\input{sec/1_intro}
\input{sec/2_Related_Work}

\input{sec/3_Method}
\input{sec/4_Experiments}
\input{sec/5_Results}
\input{sec/6_Conclusion}

\renewcommand{\acksection}{\section*{Acknowledgements}}
\begin{ack}
This work was supported by the Institute of Information \& Communications Technology Planning \& Evaluation (IITP) grant funded by the Korea government (MSIT) (IITP-2026-RS-2026-25617720, AI Star Fellowship Support Program, Contribution Rate: 40\%), and by the Culture, Sports and Tourism R\&D Program, funded by the Ministry of Culture, Sports and Tourism, through the Korea Culture Technology Planning and Evaluation Institute, an affiliated institute of the Korea Creative Content Agency in 2026 (Project Name: Development of Outpainting AI and spatial expansion technology for utilizing archive video contents, Project Number: RS-2026-25519568, Contribution Rate: 50\%), and by the Basic Science Research Program through the National Research Foundation of Korea (NRF) funded by the Ministry of Education (RS-2025-25420118, Contribution Rate: 10\%).
\end{ack}

{
    \small
    \bibliographystyle{plainnat}
    \bibliography{main}
}

\appendix
\input{sec/supp_cover}
\input{sec/X_suppl}

\newpage
\input{checklist.tex}

\end{document}

%% file: sec/0_abstract.tex
\begin{abstract}
Predicting where multiple agents will move next is fundamental to autonomous driving, robotics, and any system that must share space with people safely. Modern stochastic predictors read off inter-agent interactions once from the observed past and keep that view fixed throughout the generative rollout, so the reciprocal geometric constraints that emerge as the joint future sharpens are not re-examined. We address this gap with \textbf{SRA} (\textbf{S}patial \textbf{R}easoning \textbf{A}dapter), a modular graph denoiser that is integrated into existing diffusion and flow-matching predictors without redesigning their generative cores. At every reverse step, SRA reads the host's current future estimate, builds a sparse Euclidean interaction graph on the predicted trajectories, and returns a residual correction to the host's decoding path. Our adapter consists of three core components: a sparse neighborhood retrieved with a learned query-key score and a geometric bias, a future-conditioned relational feature recomputed on the evolving prediction, and uncertainty-weighted message passing where confident agents act as anchors for uncertain ones. We instantiate SRA on three structurally different hosts (LED, MID, and MoFlow) across four multi-agent benchmarks: SDD, NBA, Soccer, and Football. SRA improves ADE for every host on every dataset, with average gains of 5.8\% in ADE and 7.0\% in FDE. A cumulative ablation shows that the three components act as one mechanism: only together do they improve every host on every sports dataset, indicating that spatial reasoning over the model's own evolving future, weighted by how far that future can be trusted, is broadly useful across denoising architectures. Code is available at \url{https://github.com/Jaewoo97/SRA}.
\end{abstract}

%% file: sec/1_intro.tex
\section{Introduction}
\label{sec:intro}

Trajectory prediction aims to forecast the future motions of dynamic agents in a scene~\cite{alahi2016social, Xu2022RMBTP, Chib2024PTD, Pang2021TPL, Gupta2018SGAN, Fang2020TPnet, Liang2020FoP, Sun2020RLN, park2023leveraging, capellera2025jointdiff}.
As a core component of autonomous driving~\cite{luo2021safety, nayakanti2022way, phong2023truly, liu2024laformer, gu2024producing, zhao2019multi, pei2025foresight, park2024t4p, park2024improving}, robotics~\cite{salzmann2023robots}, and surveillance systems~\cite{lin2024giga}, it underpins safe motion planning in dynamic environments~\cite{lee2025interaction, lee2025nondifferentiable}.
Recent studies cast trajectory prediction as a generative problem rather than a deterministic regression task, focusing on uncertainty~\cite{tang2021collaborative, mao2023leapfrog, sun2023conformal, shi2022motion} and multi-modality in agent behaviors~\cite{liao2024cdstraj, Jeong_2025_CVPR, shi2024mtr++}. Within this view, \emph{stochastic denoisers} have emerged as the dominant recipe for multi-agent forecasting, spanning diffusion predictors such as MID~\cite{gu2022stochastic} and its leapfrog-initialized variant LED~\cite{mao2023leapfrog}, as well as flow-matching predictors such as MoFlow~\cite{fu2025moflowonestepflowmatching}.

Despite their generative strength, these models share a structural limitation: interactions among agents are encoded \emph{once}, from observed histories, and held fixed throughout the rollout~\cite{zhou2022hivt, Wong2024SoCir, bae2024singulartrajectory}. However, the interactions that decide whether a joint future is plausible are interactions between \emph{future} trajectories, such as whether two predicted paths cross or which attacker a defender ends up marking. These relations do not exist in the observed past: they appear only once the futures are predicted, and they differ from one sampled future to another. A context computed once from the past cannot represent them, so each agent's future is refined without being checked against the futures of the others. While some works inject future-aware interactions via cross-operations in a latent manifold~\cite{huang2023gameformer,fu2025moflowonestepflowmatching, jia2023towards}, such implicit coupling lacks spatial precision and offers limited interpretability for enforcing geometric relations (e.g., gaps, right-of-way, group motion)~\cite{luo2023jfp}. Rollouts therefore remain individually plausible yet globally inconsistent, exhibiting near-collisions or incoherent group behaviors that degrade prediction accuracy and downstream safety~\cite{hu2020collaborative, rowe2023fjmp, dendorfer2021mg}.

We argue that what limits these models is not how well they generate each trajectory, but how well they model the interactions among the future trajectories they generate. To this end, we introduce \textbf{S}patial \textbf{R}easoning \textbf{A}dapter (SRA), a \emph{modular graph denoiser} that models these future interactions and is integrated into an existing stochastic denoiser without redesigning its diffusion or flow-matching core. At each denoising step, SRA takes the host's own current future estimate $\hat{\mathbf{y}}$, builds a sparse agent-pair graph directly in Euclidean space, and emits a residual correction that the host injects back into its denoising update (Fig.~\ref{fig:teaser}).
Because the graph is recomputed on each sample's \emph{evolving} estimate, the $K$ modes receive $K$ different interaction structures that sharpen as the prediction does: a defender reacts to an attacker's projected run, not to the last two seconds of history. SRA is modular but not training-free: it is trained jointly with, or fine-tuned on, the host.

SRA is built around three components. First, \emph{sparse neighborhood selection}: for each agent we select $K_{\mathrm{nbr}}$ neighbors via a retrieval-style score that combines a learned semantic term with a lightweight geometric bias, so that all subsequent encoding and message passing run only over relevant neighbors. Second, \emph{agent-pair relational future encoding}: for each selected pair we build a pair-specific edge feature directly on the host's current future estimate $\hat{\mathbf{y}}^{0}$. Third, \emph{uncertainty-weighted message passing}: a predicted per-agent, per-timestep $\sigma$, calibrated via an NLL regularizer~\cite{neumeier2024reliable, capellera2025unified}, modulates routing so that confident agents act as anchors for uncertain ones.
We instantiate SRA on three hosts that span modern stochastic predictors: the leapfrog-initialized DDPM LED~\cite{mao2023leapfrog}, the vanilla DDPM MID~\cite{gu2022stochastic}, and the flow-matching model MoFlow~\cite{fu2025moflowonestepflowmatching}. We evaluate on NBA~\cite{zhan2018generating}, Soccer~\cite{scott2022soccertrack}, Football~\cite{nfl_big_data_bowl}, and SDD~\cite{robicquet2016learning}. SRA improves ADE for every host on every dataset, with average gains of 5.8\% in ADE and 7.0\% in FDE.

\noindent Our contributions are threefold:
\begin{itemize}
\item We propose \textbf{SRA}, a modular graph denoiser for iterative stochastic trajectory predictors that rebuilds an explicit interaction graph on the host's current future estimate at every reverse step and injects a residual correction without redesigning the generative core.
\item We design three components (sparse neighborhood selection, agent-pair relational future encoding, and uncertainty-weighted message passing) and show with a cumulative ablation on three sports benchmarks that they act as one mechanism: only their combination improves all nine host--dataset pairs.
\item We instantiate SRA on three structurally different hosts (LED, MID, MoFlow) and show consistent ADE/FDE improvements on four benchmarks.
\end{itemize}

%% file: sec/2_Related_Work.tex
\section{Related Work}
\label{sec:related_works}

\subsection{Trajectory prediction}
\label{sec:rw_traj}

Trajectory prediction underpins autonomous driving~\cite{zhou2023qcnet}, mobile robotics~\cite{kim2024higher}, and human-robot interaction~\cite{Gupta2018SGAN, bae2022non}. Multimodal human behavior has shifted the field from deterministic regression~\cite{Sun2020RLN, Fang2020TPnet, Pang2021TPL} to generative formulations: GANs and VAEs~\cite{Gupta2018SGAN,lee2017desire,Liang2020FoP}, then diffusion- and flow-based models~\cite{tang2021collaborative,mao2023leapfrog,liao2024cdstraj,gu2022stochastic,fu2025moflowonestepflowmatching,liu2025intdiff,yang2025unimtd}. A complementary line models inter-agent interaction via pooling~\cite{alahi2016social,Gupta2018SGAN}, attention~\cite{zhou2022hivt,ngiam2021scene, xiao2025srefiner, tang2024hpnet}, equivariant~\cite{xu2023eqmotion} and graph/hypergraph reasoning~\cite{xu2022groupnet,salzmann2023robots, thakkar2025poly, wen2024density, fang2025neuralized, jeong2024multi}, higher-order modules~\cite{socialmoif2025,kim2024higher}, and uncertainty-aware diffusion~\cite{capellera2025unified}.

A structural limitation persists across both threads: the inter-agent signal is almost always derived from \emph{observed histories} and held fixed as the predictor unrolls~\cite{zhou2022hivt, Wong2024SoCir, bae2024singulartrajectory}. SocialMOIF~\cite{socialmoif2025} distils a future-conditioned posterior into a history-only one at training time, but at inference still reduces to history cues. Scene-aware stochastic denoisers~\cite{jiang2023motiondiffuser,bae2024singulartrajectory,capellera2025unified,yang2025unimtd,liu2025intdiff} inject agent context during denoising through cross-attention over latent features rather than explicit geometric relations on the predicted futures, leaving no handle for spatial constraints such as gaps, right-of-way, or group cohesion. Methods that do condition on predicted futures do so in latent space: GameFormer~\cite{huang2023gameformer} attends to the latent futures of other agents through level-$k$ reasoning, MoFlow~\cite{fu2025moflowonestepflowmatching} mixes agents with latent attention in its decoder, and the coarse-to-fine approach~\cite{jia2023towards} refines along the temporal axis; none builds an explicit graph on the predicted coordinates or weighs a relation by the reliability of its endpoints (compared in Supp.~\ref{sec:supp_future_aware}).
To our knowledge, no prior method explicitly reasons over \emph{future-future} agent interactions in Euclidean space \emph{at inference} for scene coherence.

We close this gap with a modular graph denoising adapter on diffusion- and flow-based predictors, whose iterative denoising chain naturally supports recomputing future-trajectory interactions on the model's own evolving estimate. At every reverse step the adapter builds an explicit Euclidean interaction graph over the current future estimate and runs uncertainty-weighted message passing, injecting scene-level coherence as a residual correction without altering the host's generative core.

\subsection{Spatial reasoning with diffusion models}
\label{sec:rw_spatial_reasoning}

A recent line of work treats the denoising chain as a vehicle for \emph{structured reasoning} rather than plain sample refinement~\cite{wewer25srm}. One thread induces causal or temporal structure into the noise schedule itself: Diffusion Forcing~\cite{chen2024diffusion}, AR-Diffusion~\cite{wu2023ar}, Rolling Diffusion~\cite{pmlr-v235-ruhe24a}, and TEDi~\cite{zhang2024tedi} use position- or time-dependent schedules to enforce sequential dependencies, and SRM~\cite{wewer25srm} frames denoising as ordered structured reasoning over continuous variables.

A parallel thread injects explicit \emph{spatial} structure into diffusion: LayoutDiffusion~\cite{zheng2023layoutdiffusion}, SceneGenie~\cite{farshad2023scenegenie}, R3CD~\cite{liu2024r3cd}, and CommonScenes~\cite{zhai2023commonscenes} guide image and 3D-scene diffusion with layouts, scene-graph priors, or relational constraints. For relation prediction itself, DIFFVSGG~\cite{chen2025diffvsgg} casts video scene-graph generation as iterative denoising over object-pair embeddings, SPADE~\cite{hu2025spade} folds spatial relation reasoning into a panoptic scene-graph denoiser, and Wang et al.~\cite{wang2026demystifing} analyze how reasoning structures emerge across the video-generation chain. Across these settings, diffusion-style refinement serves as an effective spatial-reasoning mechanism when the generated entities have explicit geometric structure, particularly when reasoning is performed directly in the output domain.

Multi-agent trajectory prediction is a natural fit: each reverse step's $\hat{\mathbf{y}}^{\tau}$ already defines a temporary future scene in which Euclidean distances, relative motions, and group configurations, the quantities that govern social coherence, can be measured directly on the predicted coordinates. Our adapter accordingly treats denoising as progressive spatial reasoning: it rebuilds an explicit Euclidean interaction graph on $\hat{\mathbf{y}}^{\tau}$ at each step, runs uncertainty-weighted message passing over a sparse neighborhood, and returns a residual correction. Unlike scene-graph or layout diffusion, we operate directly in trajectory space, and in contrast to prior trajectory predictors, we enable future-conditioned interactions at inference, where ground-truth futures are unavailable, by building the graph on the model's own estimate.

\input{figures_latex/figure_method}

%% file: figures_latex/figure_method.tex
\begin{figure}
    \includegraphics[width=1.0\linewidth]{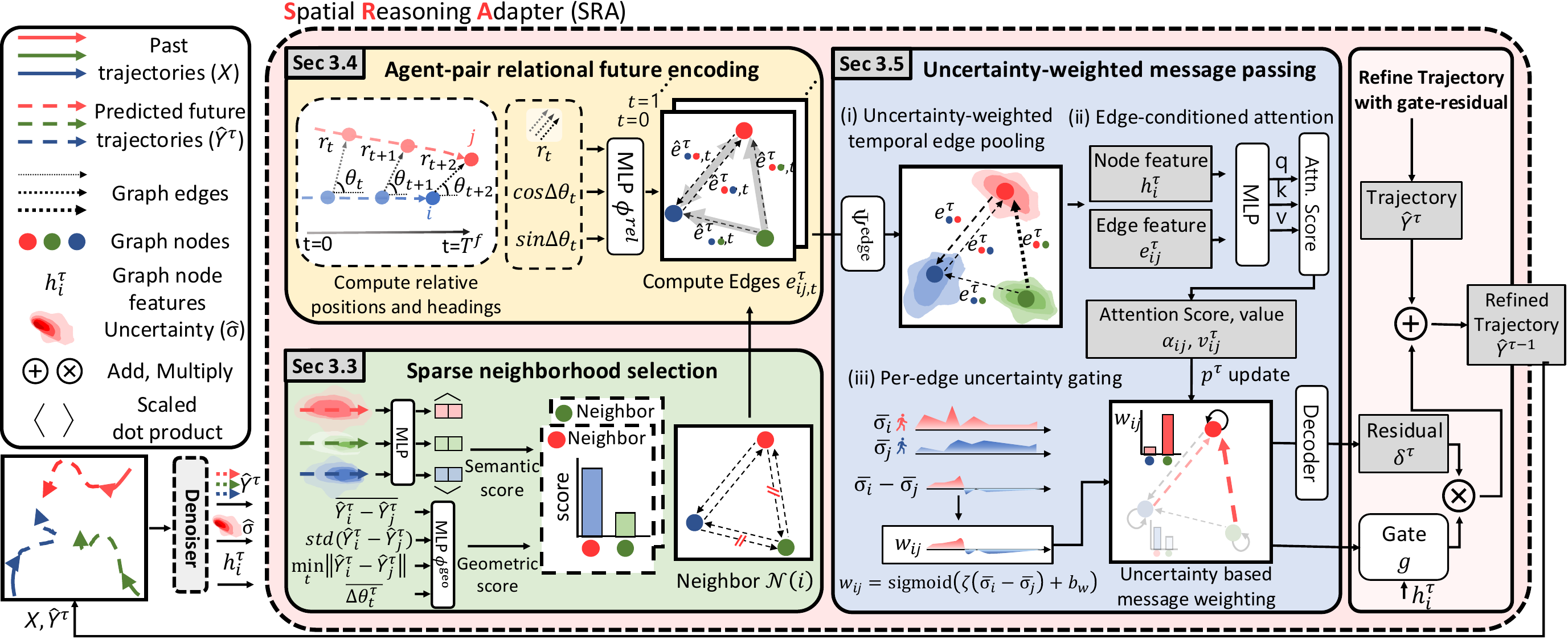} \vspace{-7pt}
    \caption{
    Overview of SRA at one reverse step $\tau$. From the host's current future estimate $\hat Y^{\tau}$, node features $h_i^{\tau}$, and uncertainty $\bar\sigma$ (bottom left), SRA \textbf{(green, Sec.~\ref{sec:sparse})} scores every pair with a semantic and a geometric term and keeps the top-$K_{\mathrm{nbr}}$ neighbors $\mathcal{N}(i)$; \textbf{(yellow, Sec.~\ref{sec:pair_encoding})} encodes the per-timestep relative positions and headings of each selected pair into edge features $\hat e^{\tau}_{ij,t}$; \textbf{(blue, Sec.~\ref{sec:unc_msg})} pools them over time with uncertainty-biased weights, uses them in edge-conditioned attention, and gates each message by $\bar\sigma_i-\bar\sigma_j$; and \textbf{(right)} decodes the aggregated message into a residual $\delta^{\tau}$ that is added through the gate $g$ to obtain the refined trajectory.
    }
    \label{fig:method}
    \vspace{-5pt}
\end{figure}

%% file: sec/3_Method.tex
\section{Method}

\subsection{Problem definition}
Trajectory forecasting aims to learn a mapping from the observed trajectories of $N$ agents
$X=\{x_{i,t}\}_{i=1,\ldots,N,\,t=T_p,\ldots,0}\in\mathbb{R}^{N\times(T_p{+}1)\times 2}$
to a set of $M$ plausible future trajectories
$
    \hat{Y}=\{\hat{Y}^{(m)}\}_{m=1}^{M}, 
    \quad
    \hat{Y}^{(m)}=\{\hat y^{0,(m)}_{i,t}\}_{i=1,\ldots,N,\,t=1,\ldots,T_f}
    \in\mathbb{R}^{N\times T_f\times 2}
$.
Following recent diffusion and flow-matching formulations, the clean future $y^{0}$ is progressively noised across levels $\tau^{k}$ (from the fully noisy $\tau^{K}$ down to the clean $\tau^{0}=0$) and the denoiser is trained to reverse that process.
In addition, we predict a per-agent, per-timestep uncertainty, parameterized as the log-variance $\mathbf{s}^{\tau}
=
\log\left(({\sigma}^{\tau})^2\right)
\in\mathbb{R}^{N\times T_f},$ which is used as a runtime signal that modulates how inter-agent messages are routed.
We write $\hat y^{\tau}$ (or $\hat y^{0}$ when the step is clear) for the host's clean-future estimate at reverse step $\tau$. All symbols are summarized in the notation table of Supp.~\ref{sec:supp_notation}.

We treat the scene as a directed interaction graph
$\mathcal{G}=(\mathcal{V},\mathcal{E})$ with $|\mathcal{V}|=N$ nodes. The neighborhood $\mathcal{N}(i)=\{j\mid(j,i)\in\mathcal{E}\}$ is \emph{dynamic}: it is rebuilt at every denoising step from the model's own current estimate of the future, so the graph's topology evolves as the prediction sharpens.

\paragraph{Fixed versus evolving interaction context.}
A history-based predictor computes its interaction context once, $c=f(X)$, and conditions every reverse step and every mode on it. (1) All $M$ modes then share the same $c$, so interactions that differ between modes (a defender engaging attacker A in one mode and attacker B in another) must be represented with mode-independent features; (2) $c$ is never revisited, so it cannot reflect the interaction structure that emerges as the futures are resolved. SRA instead builds $\mathcal{G}^{\tau,(m)}$ from each mode's current estimate $\hat Y^{\tau,(m)}$ at every reverse step, so interaction structure and trajectories sharpen together.

\subsection{Overview: a modular graph denoiser}
\label{sec:overview}

Our module takes the ``current best guess'' of the future that the host already produces, builds a sparse agent--pair graph on it, and emits a residual correction that the host injects back into its own denoising step. Concretely, at each reverse step $\tau$ the host denoiser produces
(i) a current future estimate $\hat y^{\tau}_{i}$ and
(ii) a per-agent uncertainty $\sigma^{\tau}_{i}$.
Our module ingests these and returns
$\delta^{\tau}_{i}\in\mathbb{R}^{T_f\times 2}$,
a per-agent residual that is added as a gated correction to the host's prediction:
\begin{equation}
    \tilde y^{\tau}_{i}
    =
    \hat y^{\tau}_{i}
    +
    g_i\,\delta^{\tau}_{i},
    \qquad
    g_i=\mathrm{sigmoid}\!\left(W_g\big[h_i,\bar{m}_i\big]\right),
    \label{eq:gated_residual}
\end{equation}
where $g_i\!\in\![0,1]$ is a learned scalar gate, $h_i\!\in\!\mathbb{R}^{D}$ denotes the host-provided agent feature of width $D$, $W_g$ is a learned projection, and $\bar m_i$ denotes the temporally pooled graph message produced by our module.
The refined estimate $\tilde y_i^\tau$ is passed back to the host for the reverse update.
The module is a \emph{retrieve-then-encode} pipeline of three components shared across hosts: \textbf{(a)} sparse neighborhood selection (Sec.~\ref{sec:sparse}) picks the few agents relevant to each target, \textbf{(b)} agent-pair relational future encoding (Sec.~\ref{sec:pair_encoding}) builds edge features only for them, and \textbf{(c)} uncertainty-weighted message passing (Sec.~\ref{sec:unc_msg}) aggregates messages with confidence-aware weights.
More details, including the host-specific integration, the residual decoder, and the training and inference algorithms for each host, are provided in the supplementary material (Supp.~\ref{sec:supp_impl}).

\subsection{Sparse neighborhood selection}
\label{sec:sparse}

A cheap score over all pairs selects a small relevant set, so that the expensive per-timestep encoder avoids both the quadratic cost of $N(N{-}1)$ pairs and the noise from irrelevant ones.

\paragraph{Future-conditioned graph topology.}
Instead of a topology fixed by the observed past, we build the graph on the clean future estimates $\hat y^{0}_{i}$; for $\varepsilon$-prediction hosts they are recovered from the noisy sample $y^{\tau}$ via the standard DDIM inversion
\begin{equation}
\hat y^{0}_{i}
=\bigl(y^{\tau}_{i}-\sqrt{1-\bar\alpha_{\tau}}\,\varepsilon_{\theta}(y^{\tau},x)\bigr)
\big/\sqrt{\bar\alpha_{\tau}},
\label{eq:y0_inversion}
\end{equation}

\paragraph{Two-term retrieval score.}
Based on the constructed topology, we select a sparse top-$K_{\mathrm{nbr}}$ neighborhood
$\mathcal{N}(i)$ per target $i$ using a score that combines a \emph{semantic} (learned) affinity with a \emph{geometric} (spatial) bias.
Let $r_{ij,t}^{\tau}=\hat y_{j,t}^{\tau}-\hat y_{i,t}^{\tau}$ denote the per-pair, per-timestep relative position and $\Delta\theta_{ij,t}^{\tau}$ the corresponding heading difference (formalized in Eqs.~\ref{eq:rel_pos}--\ref{eq:rel_heading}). The score is
\begin{align}
\text{score}_{ij}^{\tau}
&=\underbrace{\frac{\langle W_q\,u_i^{\tau},\,W_k\,u_j^{\tau}\rangle}{\sqrt{d_s}}}_{\text{semantic}}
+\underbrace{\phi^{\text{geo}}\!\big([\,\bar r_{ij}^{\tau},\,\operatorname{std}(r_{ij}^{\tau}),\,\min_t \|r_{ij,t}^{\tau}\|,\,\overline{\Delta\theta_{ij}^{\tau}}\,]\big)}_{\text{geometric bias}},
\label{eq:score}\\
u_i^{\tau}&=W_u\,[\hat y^{\tau}_{i,\cdot},\,\sigma^{\tau}_{i,\cdot}],
\label{eq:node_feat}
\end{align}
where $W_u$, $W_q$, and $W_k$ are learned projections, $d_s$ is the score dimension, $\bar{\cdot}$ is the temporal mean, $\operatorname{std}(\cdot)$ is the
per-axis temporal standard deviation, and $\phi^{\text{geo}}$ is an MLP.

The \textbf{semantic term} is a query--key match between \emph{what $i$ needs} ($u_i^{\tau}$: its future plan and uncertainty) and \emph{what $j$ offers} ($u_j^{\tau}$), so it can find relevance beyond proximity, e.g., two agents converging toward the same region while still far apart. The \textbf{geometric bias} supplies the explicit spatial prior it lacks through four cheap statistics of the relative trajectory: mean offset, motion variability, closest-approach distance, and mean heading difference.
The top-$K_{\mathrm{nbr}}$ candidates by $\text{score}_{ij}^{\tau}$ form $\mathcal{N}(i)$ and the sparse edge set $\mathcal{E}=\{(j,i)\mid j\in\mathcal{N}(i)\}$; we use $K_{\mathrm{nbr}}{=}5$ throughout.

\subsection{Agent-pair relational future encoding}
\label{sec:pair_encoding}

For each selected pair $(j,i)$, $j\!\in\!\mathcal{N}(i)$, we extract the relation in $i$'s agent-centric frame.
Let $R_i$ be the rotation matrix that aligns $i$'s heading at $t=0$ with the local $x$-axis, and let
$\theta_{i,t}^{0}=\operatorname{atan2}\!\big((\hat y^{0}_{i,t}-\hat y^{0}_{i,t-1})_{y},(\hat y^{0}_{i,t}-\hat y^{0}_{i,t-1})_{x}\big)$
be the instantaneous heading, where $\hat y^\tau_{i,0}$ is anchored at the last observed position $x_{i,0}$.
The per-pair, per-timestep ego-centric features are
\begin{align}
\tilde r_{ij,t}^{0}&=R_i^{\top}\!\big(\hat y^{0}_{j,t}-\hat y^{0}_{i,t}\big),
\label{eq:rel_pos}\\
\Delta\theta_{ij,t}^{0}&=\theta_{j,t}^{0}-\theta_{i,t}^{0}.
\label{eq:rel_heading}
\end{align}
$\tilde r_{ij,t}^{0}$ is the ego-centric counterpart of $r_{ij,t}^{0}$ in Eq.~\ref{eq:score}, so interactions look the same regardless of scene orientation. A small MLP $\phi_\theta^{\text{rel}}$ embeds $[\tilde r_{ij,t}^{0},\cos\Delta\theta_{ij,t}^{0},\sin\Delta\theta_{ij,t}^{0}]$ into a per-timestep future relation feature that we use as edge feature between agent nodes,
\begin{equation}
\hat e_{ij,t}^{\tau}=\phi_\theta^{\text{rel}}\!\big([\tilde r_{ij,t}^{0},\cos\Delta\theta_{ij,t}^{0},\sin\Delta\theta_{ij,t}^{0}]\big)\in\mathbb{R}^{D},
\quad t=1,\ldots,T_f,
\label{eq:edge_per_t}
\end{equation}
one $D$-dimensional feature for every selected pair at every future timestep.

\subsection{Uncertainty-weighted message passing}
\label{sec:unc_msg}

We add a log-variance head that predicts the uncertainty $\sigma^{\tau}$. A confident agent carries reliable evidence about the joint future that uncertain neighbors can lean on, so we modulate messages by uncertainty to make confident agents anchors for uncertain ones. The graph-attention layer operates on host-provided agent features $h_i^{\tau}\!\in\!\mathbb{R}^{D}$ and pooled edge features $e_{ij}^{\tau}\!\in\!\mathbb{R}^{D}$ (edge $j\!\to\!i$), and uncertainty enters at three points.

\paragraph{(i) Uncertainty-weighted temporal edge pooling.}
The per-timestep edge features $\{\hat{e}_{ij,t}^{\tau}\}_{t=1}^{T_f}$ from Eq.~\ref{eq:edge_per_t} are collapsed into a single pooled edge vector $e_{ij}^{\tau}\!\in\!\mathbb{R}^{D}$ used by message passing. The encoder $\Psi^{\text{edge}}$ first applies a temporal self-attention to mix information across the $T_f$ frames, producing $p_{ij,t}^{\tau}=\mathrm{SelfAttn}(\hat{e}_{ij,\cdot}^{\tau})_t$. A learned attention-pooling head then computes per-timestep weights, into which we inject the per-timestep relative uncertainty $\Delta\sigma_{ij,t}^{\tau}=\sigma^{\tau}_{i,t}-\sigma^{\tau}_{j,t}$ as an additive bias before softmax,
\begin{equation}
    \begin{aligned}
    p_{ij,1:T_f}^{\tau}
    &=
    \Psi^{\text{edge}}
    \left(
    \hat e_{ij,1:T_f}^{\tau}
    \right), \quad
    w_{ij,t}^{\tau}
    =
    \operatorname{softmax}_{t}
    \left(
    \phi^{\text{pool}}(p_{ij,t}^{\tau})
    +
    \Delta\sigma_{ij,t}^{\tau}
    \right), \quad
    e_{ij}^{\tau}
    =
    \sum_{t=1}^{T_f}
    w_{ij,t}^{\tau}
    p_{ij,t}^{\tau}.
    \end{aligned}
    \label{eq:edge_temporal}
\end{equation}
so frames in which the target $i$ is uncertain and the source $j$ is confident receive larger pooling weight.

\paragraph{(ii) Edge-conditioned attention.}
Queries are formed from the target node $i$ and keys/values from each source neighbor $j\!\in\!\mathcal{N}(i)$, with the future-trajectory edge feature $e_{ij}^{\tau}$ from Eq.~\ref{eq:edge_temporal} entering keys and values but not queries:
\begin{equation}
q_i^{\tau}=W_q\,h_i^{\tau}+b_q(\psi^{\tau}),
\quad
k_{ij}^{\tau}=W_k\,h_j^{\tau}+b_k(\psi^{\tau})+W_k^{(e)}e_{ij}^{\tau},
\quad
v_{ij}^{\tau}=W_v\,h_j^{\tau}+b_v(\psi^{\tau})+W_v^{(e)}e_{ij}^{\tau},
\label{eq:qkv}
\end{equation}
where $W_{\{q,k,v\}}$ and $W^{(e)}_{\{k,v\}}$ are learned projections, $b_{\{q,k,v\}}(\cdot)$ are learned biases conditioned on the noise level, and $\psi^{\tau}$ is the noise-level embedding. 

\paragraph{(iii) Per-edge uncertainty gating.}
The per-timestep $\sigma^{\tau}_{i,t}$ is too jittery to drive a per-edge gate, so we average it over the horizon, $\bar\sigma_i^{\tau}=\frac{1}{T_f}\sum_{t}\sigma_{i,t}^{\tau}$, and derive a scalar message gate
\begin{equation}
w_{ij}^{\tau}=\operatorname{sigmoid}\!\big(\zeta(\bar\sigma_i^{\tau}-\bar\sigma_j^{\tau})+b_w\big),
\label{eq:weight}
\end{equation}
where $\zeta{=}4$ is a fixed temperature and $b_w{=}0.5$ a fixed offset, so $w_{ij}^{\tau}$ grows when target $i$ is more uncertain than source $j$ and equals $\operatorname{sigmoid}(0.5)\approx0.62$ for equally confident agents.

\paragraph{Aggregation.}
Standard multi-head attention weights
$\alpha_{ij}^{\tau}=\operatorname{softmax}_{j\in\mathcal{N}(i)}
\big((q_i^{\tau})^{\top}k_{ij}^{\tau}/\sqrt{d}\big)$, with per-head dimension $d$,
are combined with the confidence gate $w_{ij}^{\tau}$ from Eq.~\ref{eq:weight} and the values $v_{ij}^{\tau}$ from Eq.~\ref{eq:qkv} into a message:
\begin{equation}
m_i^{\tau}=\sum_{j\in\mathcal{N}(i)} w_{ij}^{\tau}\,\alpha_{ij}^{\tau}\,v_{ij}^{\tau},
\qquad
h_i^{\tau,(l+1)}
=h_i^{\tau,(l)}+m_i^{\tau}+\phi_\theta^{m}\!\big(\operatorname{LN}\!\big(h_i^{\tau,(l)}+m_i^{\tau}\big)\big).
\label{eq:msg}
\end{equation}
where $\phi_\theta^{m}$ is a feed-forward block and $\operatorname{LN}$ layer normalization. After $L$ layers, a host-specific decoder $\phi^{\text{dec}}_\theta$ (Supp.~\ref{sec:supp_decoder}) maps the refined embedding to the residual $\delta_i^{\tau}$ of Eq.~\ref{eq:gated_residual}.

\subsection{Instantiation in host denoisers}
\label{sec:integration}

\paragraph{Host requirements.}
SRA assumes that the host (i) iteratively refines a continuous future estimate, (ii) exposes a prediction or intermediate representation into which a correction can be injected, and (iii) provides agent features from which an uncertainty head can be trained (it need not be native to the host). A reverse chain satisfies these naturally, since it denoises all agents' futures jointly and exposes them at every step. SRA is \emph{modular}, as it does not redesign the host's generative core, but \emph{not training-free}: it needs a host-specific injection point and is trained with the host (Sec.~\ref{sec:training}). Regression- and classification-based predictors are left to future work.

\paragraph{Hosts.} We use three hosts:
\textbf{MID}~\cite{gu2022stochastic}, a vanilla DDPM with a pure transformer denoiser and no built-in inter-agent reasoning;
\textbf{LED}~\cite{mao2023leapfrog}, a leapfrog-initialized DDPM that collapses the reverse chain to five fine-grained $\varepsilon$-prediction steps;
and \textbf{MoFlow}~\cite{fu2025moflowonestepflowmatching}, a flow-matching model whose Motion-Transformer decoder directly regresses clean trajectories with factorized $K$-attention (modes) and $A$-attention (agents).
Implementation details for each host are in Supp.~\ref{sec:supp_impl}.

\subsection{Training}
\label{sec:training}

SRA leaves the host's generative objective $\mathcal{L}_{\text{gen}}$ (e.g., DDPM $\varepsilon$-MSE or flow matching) unchanged and only adds a single heteroscedastic NLL term that trains the log-variance head alongside it.
Given the host's prediction residual $r_\theta$ and its predicted variance $\sigma_\theta^{2}$, we minimize
\begin{equation}
\mathcal{L}_{\sigma}
=\tfrac{1}{2}\mathbb{E}\!\left[\log\sigma_\theta^{2}+\|r_\theta\|^{2}/\sigma_\theta^{2}\right],
\label{eq:nll}
\end{equation}
where $r_\theta$ is the noise residual $\varepsilon-\varepsilon_\theta$ for noise-prediction hosts or the trajectory residual $y^{0}-\hat y^{0}_\theta$ for flow-matching hosts. The total objective is $\mathcal{L}_{\text{total}}=\mathcal{L}_{\text{gen}} + \lambda_\sigma\,\mathcal{L}_{\sigma}$ with $\lambda_\sigma{=}10^{-2}$. We detach $r_\theta$ from the NLL gradient path so $\mathcal{L}_\sigma$ trains only the log-variance head and leaves the host's denoising parameters untouched.

\input{tables/main_quant}

%% file: tables/main_quant.tex
\newcommand{\gain}[1]{{\scriptsize\textcolor{ForestGreen}{(+#1\%)}}}
\newcommand{\loss}[1]{{\scriptsize($-$#1\%)}}

\begin{table*}[t]
\caption{\textbf{Main results.} ADE / FDE (best of $K{=}20$ modes; lower is better) of three hosts with and without SRA on NBA, Soccer, Football, and SDD. Percentages are the relative improvement of SRA over its host; the only negative value (MID, FDE on SDD) is shown in black.}
\label{main_sports}
\centering
\setlength{\tabcolsep}{3.5pt}
\resizebox{\textwidth}{!}{%
\begin{tabular}{l cc cc cc cc}
\toprule
& \multicolumn{2}{c}{NBA} & \multicolumn{2}{c}{Soccer} & \multicolumn{2}{c}{Football} & \multicolumn{2}{c}{SDD} \\
\cmidrule(lr){2-3} \cmidrule(lr){4-5} \cmidrule(lr){6-7} \cmidrule(lr){8-9}
Method & ADE & FDE & ADE & FDE & ADE & FDE & ADE & FDE \\
\midrule
MID~\cite{gu2022stochastic}          & 1.037 & 1.472 & 0.451 & 0.783 & 0.274 & 0.459 & 8.85 & 15.92 \\
MID + SRA    & 0.957\,\gain{7.7} & 1.346\,\gain{8.6} & 0.370\,\gain{18.0} & 0.595\,\gain{24.0} & 0.265\,\gain{3.3} & 0.438\,\gain{4.6} & 8.53\,\gain{3.6} & 16.61\,\loss{4.3} \\
\midrule
LED~\cite{mao2023leapfrog}          & 0.839 & 1.204 & 0.372 & 0.821 & 0.289 & 0.580 & 8.14 & 13.95 \\
LED + SRA    & 0.778\,\gain{7.3} & 1.161\,\gain{3.6} & 0.364\,\gain{2.2} & 0.710\,\gain{13.5} & 0.253\,\gain{12.5} & 0.505\,\gain{12.9} & 7.91\,\gain{2.8} & 13.31\,\gain{4.6} \\
\midrule
MoFlow~\cite{fu2025moflowonestepflowmatching}       & 0.703 & 0.884 & 0.406 & 0.593 & 0.251 & 0.421 & 5.44 & 8.30 \\
MoFlow + SRA & 0.695\,\gain{1.1} & 0.870\,\gain{1.6} & 0.374\,\gain{7.9} & 0.522\,\gain{12.0} & 0.245\,\gain{2.4} & 0.412\,\gain{2.1} & 5.37\,\gain{1.3} & 8.21\,\gain{1.1} \\
\bottomrule
\end{tabular}%
}
\end{table*}

%% file: sec/4_Experiments.tex
\section{Experiments}
\subsection{Dataset}
We use 4 datasets to evaluate the prediction performance: SDD~\cite{robicquet2016learning}, NBA~\cite{zhan2018generating}, Soccer~\cite{scott2022soccertrack}, Football~\cite{nfl_big_data_bowl}.
While SDD is widely adopted as a benchmark for trajectory prediction, its patterns are monotonous as it only includes pedestrians walking.
On the other hand, sports datasets (NBA, Soccer, Football) consist of dynamic motion and interaction, greatly increasing the difficulty of predicting future motion.
Therefore, we focus on the sports datasets: the component analysis is run on all three sports datasets, and the finer-grained ablations on NBA, the most demanding of them, where the court is small relative to the number of agents, nearly every player is within interaction range of several others, and configurations turn over quickly across possessions.
For sports datasets, we vary the frame rate (5Hz for NBA, 5Hz for football, 3.33Hz for soccer) with 10 past frames to predict 20 frames. For SDD, we follow the standard protocol of 8 past frames (3.2\,s) and 12 future frames (4.8\,s) at 2.5\,Hz.

\subsection{Metrics}
We report Average Displacement Error (ADE) and Final Displacement Error (FDE), computed as the minimum over $K{=}20$ predicted modes following the standard protocol of prior work~\cite{mao2023leapfrog, gu2022stochastic, fu2025moflowonestepflowmatching}.
We do not use collision rate: physical contact between players is a normal part of basketball, soccer, and football, so a collision counter would penalize predictions that are in fact correct.

%% file: sec/5_Results.tex
\section{Results}

\input{figures_latex/figure_qual_combined}

\subsection{Quantitative results}
Table~\ref{main_sports} compares SRA against generative baselines on sports (NBA, Soccer, Football) and pedestrian (SDD) benchmarks.
SRA improves ADE for every host on every dataset and FDE in 11 of 12 host--dataset pairs, by 5.8\% ADE and 7.0\% FDE on average, with the largest gains on sports scenes: up to 18.0\% ADE and 24.0\% FDE over the host (MID on Soccer).
Sports scenes involve dense, fast, ball-coupled interactions where future-conditioned coherence matters most, whereas pedestrian dynamics are comparatively monotone and leave less headroom for graph corrections.

Further experiments are provided in the supplementary material: an analysis with the joint metrics JADE/JFDE~\cite{weng2023joint} (Supp.~\ref{sec:supp_joint}), a comparison with future-aware interaction modules (Supp.~\ref{sec:supp_future_aware}), and the inference latency of SRA (Supp.~\ref{sec:supp_latency}).

\subsection{Qualitative results}
Figure~\ref{fig:qual_main} compares SRA's predictions against the baseline on three representative NBA scenes.
In the first row (LED), SRA correctly predicts the defender's return-to-defense motion that the baseline misses.
In the second row (MID), SRA captures a tight ball-carrier coupling, so the ball's trajectory tracks its carrier rather than drifting independently.
In the third row (MoFlow), SRA produces an accurate offensive sequence in which the ball is passed cleanly along the attacking players.

\subsection{Ablations}
\label{sec:ablations}
\input{tables/ablation_cumulative}

\paragraph{Cumulative component analysis.}
Table~\ref{tab:ablation_cumulative} adds the three components one at a time on all three sports datasets.
Leveraging interaction from the predicted future is what produces the gain: relational encoding on a dense graph already improves ADE in 6 of the 9 host--dataset pairs.
Using that signal well, however, requires the other two components. Sparsifying without uncertainty is close to a coin flip (better than the dense graph in 4 pairs and worse in 5) because top-$K_{\mathrm{nbr}}$ selection commits hard to a still-noisy estimate, and a wrong choice discards the neighbors that mattered.
Uncertainty tells the module how far to trust that estimate, and only with all three components does SRA improve all nine pairs in both metrics.
The components are therefore one mechanism rather than independent additions: the future interaction signal, the selection that concentrates it, and the confidence estimate that makes the selection safe.
This also explains why ablating a single design choice of the full model (Tables~\ref{tab:ablation_interaction_features} and~\ref{tab:ablation_neighbor_selection}) changes the result by a smaller margin than the full gain over the host: it measures a broken dependency rather than the size of an independent contribution.

\paragraph{Use of uncertainty.}
\input{figures_latex/figure_ablation_uncertainty}
The last two rows of each block in Table~\ref{tab:ablation_cumulative} isolate uncertainty-weighted message passing, which improves ADE and FDE for every host on every sports dataset, with the largest relative ADE reductions for MID on Soccer ($0.455\!\rightarrow\!0.370$, 18.7\%) and LED on Football ($0.306\!\rightarrow\!0.253$, 17.3\%).
Each message is modulated by the \emph{difference} $\sigma_i-\sigma_j$ (Eqs.~\ref{eq:edge_temporal} and~\ref{eq:weight}), so what matters is which of the two agents the host currently localizes better: the more certain agent in each pair becomes the anchor that the uncertain one leans on, and relations propagate in the direction in which they can be trusted.
Figure~\ref{fig:uncertainty_comparison} stratifies NBA scenes into five bins by predicted $\sigma$. The LED baseline ADE grows monotonically with $\sigma$, so $\sigma$ is well-calibrated; the improvement of LED + SRA appears in every bin and rises with $\sigma$, reaching more than $2\times$ the lowest-bin gain in the highest bin, so SRA concentrates its corrections where the host is least reliable.
Figure~\ref{fig:qual_uncertainty} shows the effect qualitatively: SRA assigns the highest $\sigma$ to the ball carrier and its defender, the two most coupled agents, and routes messages from more confident players to them, so the predicted paths follow the actual pass to the corner; without uncertainty (left), the ball drifts into an implausible trajectory that misses the pass.

\input{tables/ablation_feature_neighbor}
\paragraph{Relational features.}
Table~\ref{tab:ablation_interaction_features} compares four configurations of pairwise relational features used in the future relational feature $\hat e_{ij,t}^{\tau}$.
Using only L2 distance, the simplest geometric cue, already yields gains over the base denoisers, confirming that \emph{any} explicit future-trajectory interaction improves over latent-only attention.
Relative position adds directional information and improves FDE in MID and LED, while heading direction alone provides comparable ADE but stronger FDE gains in MoFlow.
Combining relative position with heading direction (our default) achieves the best balance across all three backbones, as it captures both where agents are heading and how they are oriented relative to each other.

\paragraph{Neighbor selection method.}
Table~\ref{tab:ablation_neighbor_selection} compares our learned two-term retrieval score (Eq.~\ref{eq:score}) against a rule-based selection of the $K_{\mathrm{nbr}}$ agents closest in mean future L2 distance, and against each of its two terms alone.
Either learned term alone already improves FDE over the rule-based selection for every host, and the two perform almost identically, but their combination is best in every cell.
The terms are complementary: the geometric bias supplies a proximity prior that the semantic term has no explicit access to, while the semantic query--key match captures relevance that distance misses, such as two agents converging on the same region while still far apart.
The gaps are modest, indicating that the graph is fairly robust to the selection rule.

%% file: figures_latex/figure_qual_combined.tex
\begin{figure}[t]
    \centering
    \begin{minipage}[t]{0.6\linewidth}
        \vspace{0pt}\centering
        \includegraphics[width=\linewidth]{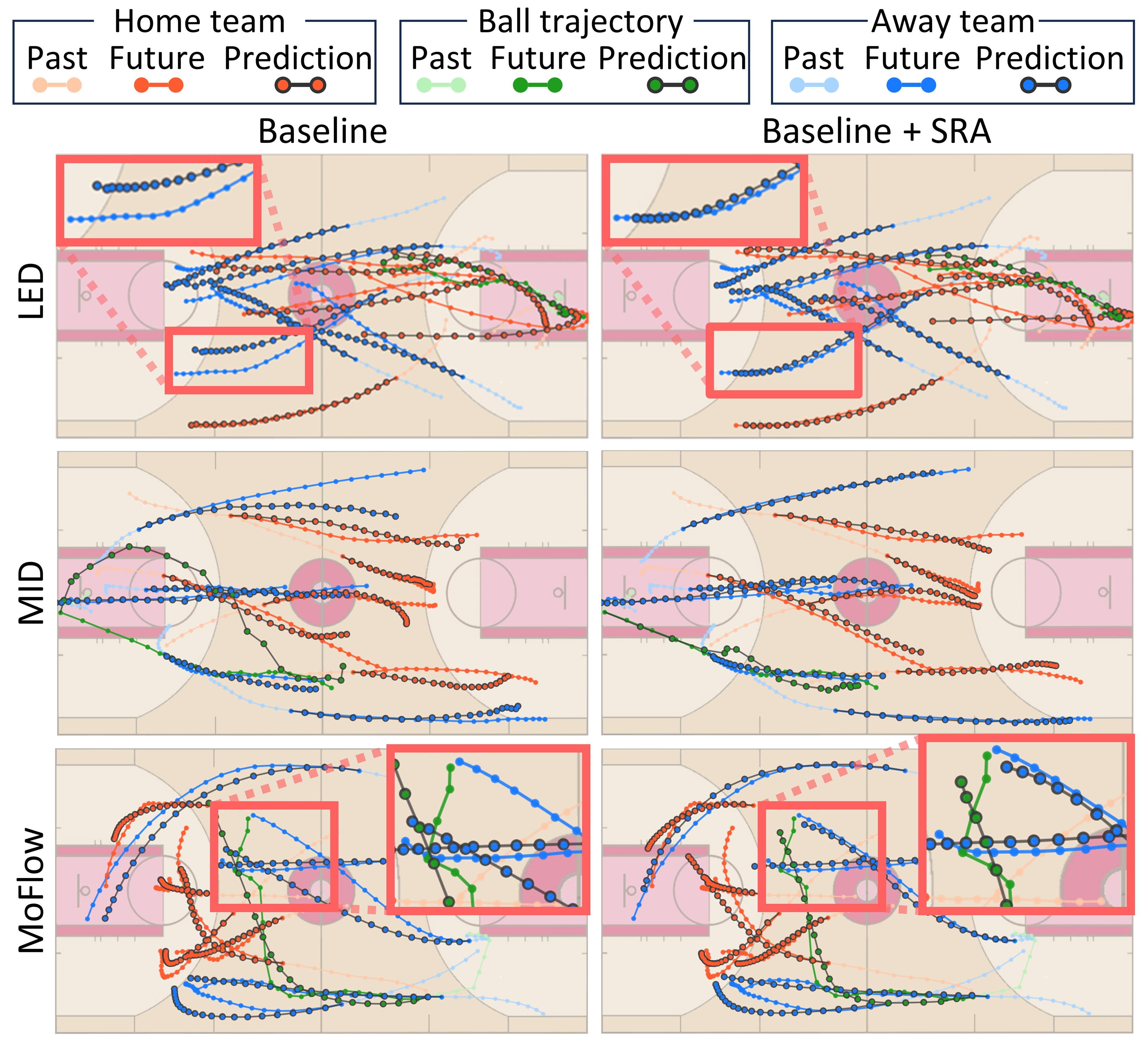}
        \vspace{-16pt}
        \caption{Qualitative comparison on NBA. Rows: LED, MID, MoFlow; left: host, right: host + SRA. Orange, green, and blue are home players, the ball, and away players; light, medium, and dark dots are the observed past, the ground-truth future, and the prediction (best of $K{=}20$ modes by ADE). Red boxes mark where the predictions differ.}
        \label{fig:qual_main}
    \end{minipage}\hfill
    \begin{minipage}[t]{0.36\linewidth}
        \vspace{0pt}\centering
        \includegraphics[width=\linewidth]{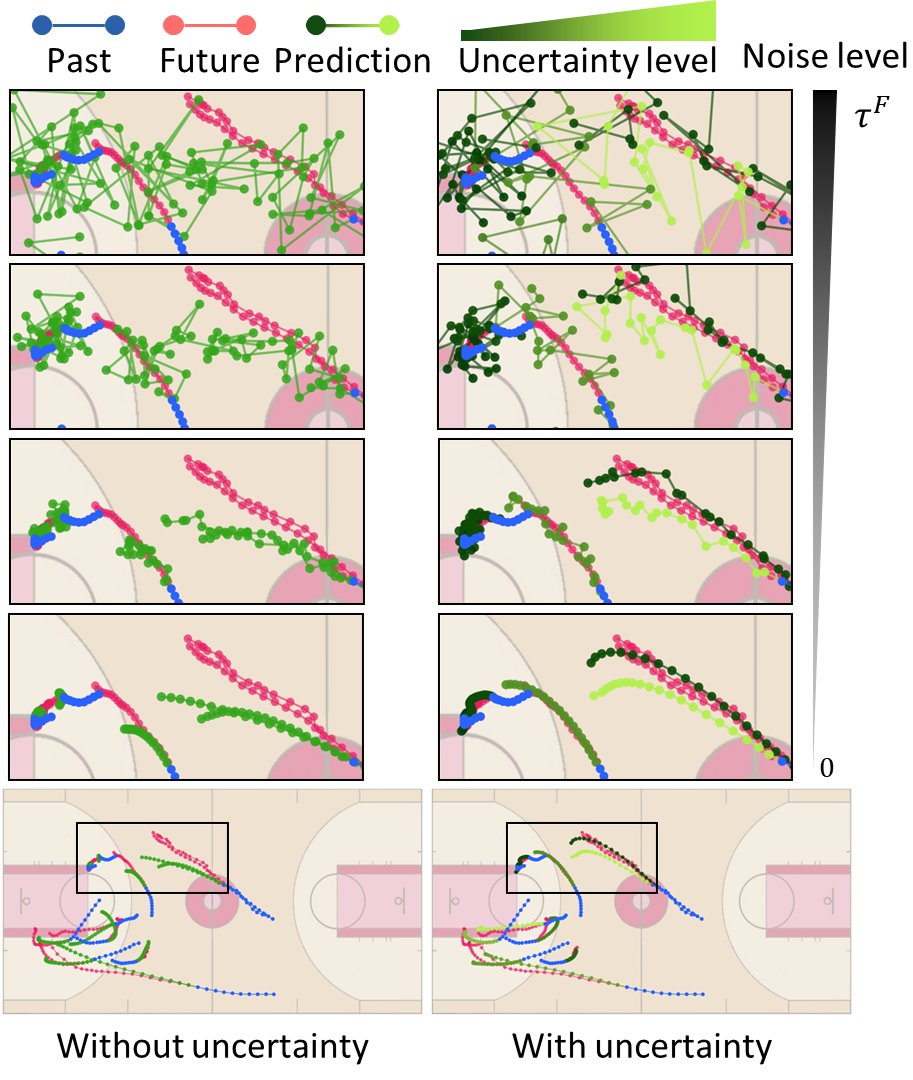}
        \vspace{-16pt}
        \caption{Denoising without (left) and with (right) uncertainty, LED on NBA. Bottom: full scene; above: the boxed region from high noise (top) to the final step. Blue: past; pink: ground truth; green: prediction, shaded by predicted $\sigma$ (right).}
        \label{fig:qual_uncertainty}
    \end{minipage}
    \vspace{-8pt}
\end{figure}

%% file: tables/ablation_cumulative.tex
\begin{table}[t]
\centering
\caption{\textbf{Cumulative component analysis} (ADE / FDE). Starting from the unmodified host, we add one SRA component at a time: future relational encoding over all neighbors (Enc.), sparse top-$K_{\mathrm{nbr}}$ neighbor selection (Sel.), and uncertainty-weighted message passing (Unc.). The last row of each block is the full SRA reported in Table~\ref{main_sports}. Best per dataset and host in \textbf{bold}.}
\label{tab:ablation_cumulative}
\scriptsize
\setlength{\tabcolsep}{4pt}
\resizebox{0.92\linewidth}{!}{%
\begin{tabular}{@{}ll ccc ccc@{}}
\toprule
Dataset & Configuration & Enc. & Sel. & Unc. & MID & LED & MoFlow \\
\midrule
\multirow{4}{*}{NBA}
 & Host baseline              &            &            &            & 1.037 / 1.472 & 0.839 / 1.204 & 0.703 / 0.884 \\
 & + encoding (all neighbors) & \checkmark &            &            & 1.005 / 1.417 & \textbf{0.776} / 1.176 & 0.698 / 0.876 \\
 & + sparse selection         & \checkmark & \checkmark &            & 0.963 / 1.367 & 0.886 / 1.238 & 0.705 / 0.881 \\
 & + uncertainty (full SRA)   & \checkmark & \checkmark & \checkmark & \textbf{0.957} / \textbf{1.346} & 0.778 / \textbf{1.161} & \textbf{0.695} / \textbf{0.870} \\
\midrule
\multirow{4}{*}{Soccer}
 & Host baseline              &            &            &            & 0.451 / 0.783 & 0.372 / 0.821 & 0.406 / 0.593 \\
 & + encoding (all neighbors) & \checkmark &            &            & 0.438 / 0.648 & 0.384 / 0.748 & 0.386 / \textbf{0.520} \\
 & + sparse selection         & \checkmark & \checkmark &            & 0.455 / 0.736 & 0.383 / 0.822 & 0.382 / 0.533 \\
 & + uncertainty (full SRA)   & \checkmark & \checkmark & \checkmark & \textbf{0.370} / \textbf{0.595} & \textbf{0.364} / \textbf{0.710} & \textbf{0.374} / 0.522 \\
\midrule
\multirow{4}{*}{Football}
 & Host baseline              &            &            &            & 0.274 / 0.459 & 0.289 / 0.580 & 0.251 / 0.421 \\
 & + encoding (all neighbors) & \checkmark &            &            & 0.269 / 0.454 & 0.305 / 0.682 & 0.251 / \textbf{0.395} \\
 & + sparse selection         & \checkmark & \checkmark &            & 0.296 / 0.475 & 0.306 / 0.643 & 0.247 / 0.422 \\
 & + uncertainty (full SRA)   & \checkmark & \checkmark & \checkmark & \textbf{0.265} / \textbf{0.438} & \textbf{0.253} / \textbf{0.505} & \textbf{0.245} / 0.412 \\
\bottomrule
\end{tabular}%
}
\vspace{-6pt}
\end{table}

%% file: figures_latex/figure_ablation_uncertainty.tex
\begin{figure}[t]
    \centering
    \begin{minipage}[c]{0.44\linewidth}
        \includegraphics[width=\linewidth]{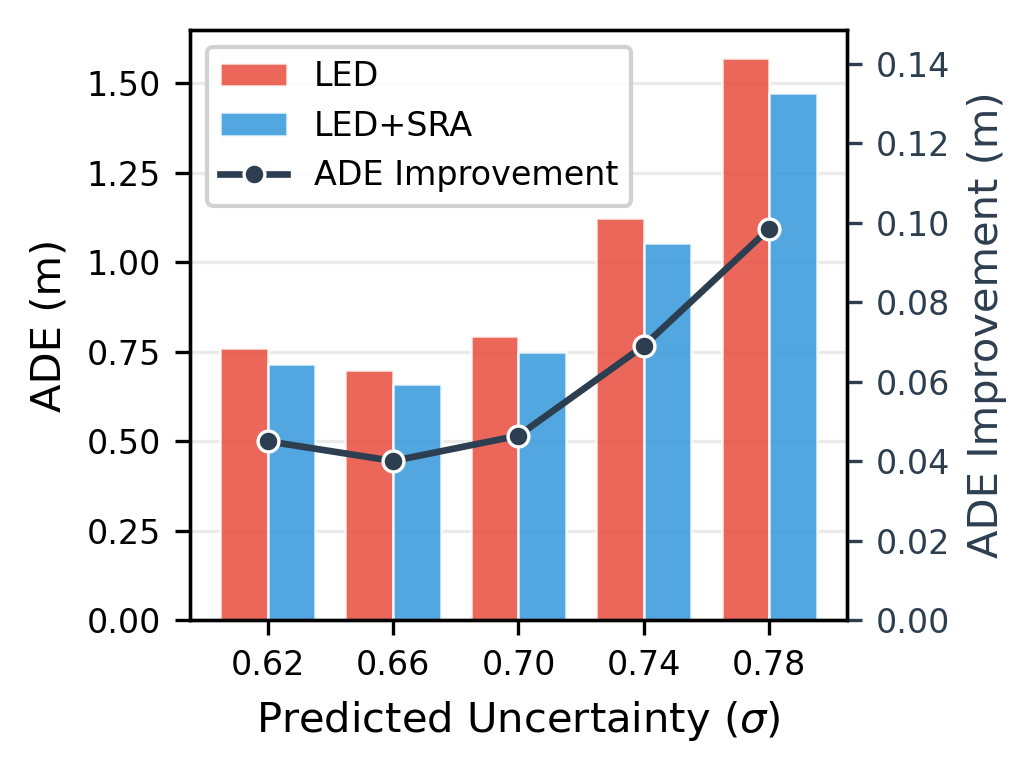}
    \end{minipage}\hfill
    \begin{minipage}[c]{0.53\linewidth}
    \caption{
    ADE of LED vs.\ LED+SRA stratified by predicted uncertainty $\sigma$ on the NBA dataset (bars, left axis), and the ADE improvement of LED+SRA over LED (line, right axis). The baseline error grows with $\sigma$, and SRA's gain appears in every bin and is largest in the highest-$\sigma$ bin.
    }
    \label{fig:uncertainty_comparison}
    \end{minipage}
    \vspace{-8pt}
\end{figure}

%% file: tables/ablation_feature_neighbor.tex
\begin{table}[t]
\centering
\begin{minipage}[t]{0.49\linewidth}
\centering
\scriptsize
\setlength{\tabcolsep}{2.2pt}
\begin{tabular}{@{}ccc cc cc cc@{}}
\toprule
\multicolumn{3}{c}{Relational feat.} & \multicolumn{2}{c}{MID} & \multicolumn{2}{c}{LED} & \multicolumn{2}{c}{MoFlow} \\
\cmidrule(lr){1-3} \cmidrule(lr){4-5} \cmidrule(lr){6-7} \cmidrule(lr){8-9}
L2 & Pos. & Head. & ADE & FDE & ADE & FDE & ADE & FDE \\
\midrule
\checkmark &            &            & 0.968 & 1.376 & 0.781 & 1.164 & 0.699 & 0.878 \\
           & \checkmark &            & 0.966 & 1.360 & \textbf{0.778} & \textbf{1.153} & 0.698 & 0.880 \\
           &            & \checkmark & 0.967 & 1.353 & 0.784 & 1.163 & 0.698 & 0.872 \\
           & \checkmark & \checkmark & \textbf{0.957} & \textbf{1.346} & \textbf{0.778} & 1.161 & \textbf{0.695} & \textbf{0.870} \\
\bottomrule
\end{tabular}
\vspace{2pt}
\caption{\textbf{Relational features} used in $\hat e_{ij,t}^{\tau}$ (Eq.~\ref{eq:edge_per_t}) on NBA: L2 distance (L2), relative position (Pos.), heading difference (Head.), or relative position + heading (ours, last row). ADE / FDE, best of 20; best per column in \textbf{bold}.}
\label{tab:ablation_interaction_features}
\end{minipage}\hfill
\begin{minipage}[t]{0.49\linewidth}
\centering
\scriptsize
\setlength{\tabcolsep}{2.2pt}
\begin{tabular}{@{}ccc cc cc cc@{}}
\toprule
Rule-based & \multicolumn{2}{c}{Ours} & \multicolumn{2}{c}{MID} & \multicolumn{2}{c}{LED} & \multicolumn{2}{c}{MoFlow} \\
\cmidrule(lr){1-1} \cmidrule(lr){2-3} \cmidrule(lr){4-5} \cmidrule(lr){6-7} \cmidrule(lr){8-9}
L2 & Geo. & Sem. & ADE & FDE & ADE & FDE & ADE & FDE \\
\midrule
\checkmark &            &            & 0.963 & 1.367 & 0.780 & 1.182 & 0.701 & 0.881 \\
           & \checkmark &            & 0.959 & 1.348 & 0.782 & 1.172 & 0.697 & 0.873 \\
           &            & \checkmark & 0.960 & 1.351 & 0.781 & 1.169 & 0.696 & 0.872 \\
           & \checkmark & \checkmark & \textbf{0.957} & \textbf{1.346} & \textbf{0.778} & \textbf{1.161} & \textbf{0.695} & \textbf{0.870} \\
\bottomrule
\end{tabular}
\vspace{2pt}
\caption{\textbf{Neighbor scoring} on NBA. Rule-based: the $K_{\mathrm{nbr}}$ closest agents by mean future L2 distance (L2). Ours: the geometric (Geo.) and semantic (Sem.) terms of our learned retrieval score (Eq.~\ref{eq:score}), used alone or together (full SRA, last row). ADE / FDE, best of 20.}
\label{tab:ablation_neighbor_selection}
\end{minipage}
\vspace{-6pt}
\end{table}

%% file: sec/6_Conclusion.tex
\section{Conclusion}
\label{sec:conclusion}

We presented SRA, a modular graph denoiser that is integrated into iterative stochastic trajectory predictors and reasons about inter-agent interaction directly in the Euclidean future trajectory space.
At every reverse step, SRA rebuilds an explicit interaction graph on the host's current future estimate, runs uncertainty-weighted message passing over a sparse $K_{\mathrm{nbr}}$-neighborhood, and emits a residual correction that the host injects back into its denoising update without redesigning the generative core.
Our cumulative ablation shows that its three components (sparse neighborhood selection, agent-pair relational future encoding, and uncertainty-weighted message passing) work as one mechanism, and only their combination improves every host on every sports dataset.
Instantiated on three structurally different hosts (MID, LED, MoFlow), SRA improves every host on every dataset across four multi-agent benchmarks (NBA, Soccer, Football, SDD) in ADE, at a cost of a few milliseconds per scene.

\paragraph{Limitations.}
SRA is modular but not training-free: it requires a host-specific injection point and joint training or fine-tuning with the host. It assumes an iterative host that refines continuous future estimates, so extending it to regression- or classification-based predictors remains future work. Its graph is built on the host's own estimate and inherits that estimate's errors, and our evaluation does not cover mixed-class scenes with map-dependent interactions. Supp.~\ref{sec:supp_limitations} discusses these points in detail.

%% file: sec/supp_cover.tex
\clearpage
\vbox{%
  \hsize\textwidth
  \linewidth\hsize
  \vskip 0.1in
  \hrule height 4pt
  \vskip 0.25in
  \centering
  {\LARGE\bf SRA: Spatial Reasoning Adapter via Evolving Social\\ Interaction Graphs for Trajectory Prediction\par}
  \vskip 0.15in
  {\Large Supplementary Material\par}
  \vskip 0.25in
  \hrule height 1pt
  \vskip 0.09in
}

\vspace{1.5em}
\noindent This supplementary material accompanies the main paper. It provides the notation used throughout, full implementation details of SRA in each host, additional ablations and analyses, and more qualitative results. Sections labeled with letters (Supp.~A--I) are in this document; numbered sections refer to the main paper.

\vspace{1.5em}
\noindent{\large\bf Contents}
\vspace{0.6em}

\newcommand{\suppentry}[2]{%
  \noindent\hyperref[#1]{\makebox[1.8em][l]{\ref*{#1}}#2}\dotfill\pageref{#1}\par\vspace{4pt}}

{\setlength{\parskip}{0pt}
\suppentry{sec:supp_notation}{Notation}
\suppentry{sec:supp_impl}{Implementation Details}
\suppentry{sec:supp_ablations}{Additional Ablations}
\suppentry{sec:supp_joint}{Joint Metrics}
\suppentry{sec:supp_future_aware}{Comparison with Future-Aware Interaction Modules}
\suppentry{sec:supp_robustness}{Robustness Across Scenes}
\suppentry{sec:supp_latency}{Inference Latency}
\suppentry{sec:supp_limitations}{Limitations}
\suppentry{sec:more_qual}{More Qualitative Results}
}

\clearpage

%% file: sec/X_suppl.tex
\section{Notation}
\label{sec:supp_notation}

\begin{table}[!ht]
\centering
\scriptsize
\setlength{\tabcolsep}{4pt}
\caption{\textbf{Notation} used in Sec.~3 of the main paper.}
\label{tab:notation}
\begin{tabular}{@{}ll@{}}
\toprule
Symbol & Meaning \\
\midrule
$N$, $T_p$, $T_f$ & number of agents; observed and predicted horizon lengths \\
$X$, $x_{i,t}$ & observed trajectories; position of agent $i$ at past step $t$ ($t{=}0$: last observation) \\
$M$ ($=K{=}20$) & number of sampled future modes \\
$\tau$, $\tau^{k}$ & noise level / reverse step ($\tau^{K}$: pure noise, $\tau^{0}{=}0$: clean) \\
$y^{\tau}$ & noisy future sample at level $\tau$ \\
$\hat y^{0}$, $\hat y^{\tau}$ & host's current clean-future estimate at reverse step $\tau$ \\
$\tilde y^{\tau}_{i}$ & estimate after SRA's gated residual correction (Eq.~\ref{eq:gated_residual}) \\
$\delta^{\tau}_{i}\in\mathbb{R}^{T_f\times 2}$ & SRA residual for agent $i$ \\
$g_i$ & learned gate on the residual \\
$h_i^{\tau}\in\mathbb{R}^{D}$ & host-provided agent feature (node feature of the graph) \\
$\sigma^{\tau}_{i,t}$, $\mathbf{s}^{\tau}$ & predicted uncertainty of agent $i$ at future step $t$; its log-variance \\
$\bar\sigma^{\tau}_{i}$ & temporal mean of $\sigma^{\tau}_{i,t}$ (Sec.~\ref{sec:unc_msg}) \\
$\mathcal{G}=(\mathcal{V},\mathcal{E})$, $\mathcal{N}(i)$ & directed interaction graph; selected neighbors of target $i$ (edges $j\!\to\!i$) \\
$K_{\mathrm{nbr}}$ & number of selected neighbors per agent ($=5$) \\
$u_i^{\tau}$ & node feature used by the retrieval score (Eq.~\ref{eq:node_feat}) \\
$r^{\tau}_{ij,t}$, $\tilde r^{0}_{ij,t}$ & relative position of $j$ w.r.t.\ $i$; the same in $i$'s ego-centric frame \\
$\theta^{0}_{i,t}$, $\Delta\theta^{0}_{ij,t}$ & heading of $i$; heading difference between $j$ and $i$ \\
$R_i$ & rotation aligning $i$'s heading at $t{=}0$ with the $x$-axis \\
$\hat e^{\tau}_{ij,t}$ & per-timestep future relational feature of edge $j\!\to\!i$ (Eq.~\ref{eq:edge_per_t}) \\
$p^{\tau}_{ij,t}$, $e^{\tau}_{ij}$ & temporally mixed edge feature; uncertainty-pooled edge feature (Eq.~\ref{eq:edge_temporal}) \\
$w^{\tau}_{ij,t}$ & temporal pooling weight biased by $\Delta\sigma^{\tau}_{ij,t}=\sigma^{\tau}_{i,t}-\sigma^{\tau}_{j,t}$ \\
$w^{\tau}_{ij}$, $\zeta$, $b_w$ & per-edge uncertainty gate; its temperature ($4$) and offset ($0.5$) (Eq.~\ref{eq:weight}) \\
$\alpha^{\tau}_{ij}$, $m^{\tau}_i$ & attention weight; aggregated message (Eq.~\ref{eq:msg}) \\
$\psi^{\tau}$ & noise-level embedding \\
$L$ & number of message-passing layers \\
$\lambda_\sigma$ & weight of the NLL loss $\mathcal{L}_\sigma$ ($10^{-2}$) \\
\bottomrule
\end{tabular}
\end{table}

\section{Implementation Details}
\label{sec:supp_impl}

\subsection{Graph Integration: Output-Level vs.\ Embedding-Level}
\label{sec:supp_injection}

Although SRA's graph module exposes the same interface ($\hat{y}^{\tau},\,x,\,\tau,\,\sigma$) across all three host denoisers, the \emph{point of injection} differs between MID/LED (output-level) and MoFlow (embedding-level). We explain the architectural motivation for each choice below.

\paragraph{Output-level injection in MID and LED.}
MID's vanilla transformer denoiser produces a noise estimate $\varepsilon_\theta$ directly from a sequence of past-context tokens; it has no intermediate layer that forms explicit agent representations of \emph{future} trajectory state. LED's core denoiser is similarly structured, and its weights are \emph{frozen} during SRA fine-tuning, which makes internal surgery impractical. Because neither architecture exposes a natural ``future agent embedding'' between encoder and decoder, we inject SRA's correction at the only available boundary: the final output.

Concretely, SRA maps the host's current $\hat{y}^{\tau}$ through the graph module to a residual $\delta^{\tau}\!\in\!\mathbb{R}^{A\times K\times T_f\times 2}$, then modifies the noise prediction directly:
\begin{equation}
\varepsilon_\theta' = \varepsilon_\theta + g_i\,\delta^{\tau}_i,
\label{eq:output_injection}
\end{equation}
where $g_i\!\in\![0,1]$ is the learned scalar gate of Eq.~\ref{eq:gated_residual}. Because the correction operates in $\varepsilon$-space, the subsequent DDPM reverse step naturally translates it into a geometric shift on the predicted clean trajectory, with the correct noise-schedule scaling built in.

\paragraph{Embedding-level injection in MoFlow.}
MoFlow's Motion-Transformer-style decoder factorizes attention across trajectory modes ($K$-attention) and across agents ($A$-attention) in two sequential blocks. This provides a natural intermediate representation: after $K$-attention has already mixed information across modes, but before $A$-attention fuses inter-agent context, the node embedding matrix $\mathbf{Y}_\mathrm{emb}\!\in\!\mathbb{R}^{B\times K\times A\times D}$ captures per-agent, per-mode state. Injecting here is strictly more expressive than output-level: the graph-refined embeddings then propagate through $A$-attention (whose inter-agent queries and keys depend on them), the context-fusion MLP, the motion decoder, and the regression head, giving the full decoder capacity to interpret the graph signal rather than bypassing the decoder with a simple additive correction.

The injection is therefore:
\begin{equation}
\mathbf{Y}_\mathrm{emb} \leftarrow \mathrm{FutureGraph}\!\left(\mathbf{Y}_\mathrm{emb},\; \hat{y}^{\tau}_\mathrm{abs},\; \mathbf{t}_\mathrm{emb},\; \tau,\; \sigma\right),
\label{eq:emb_injection}
\end{equation}
after which $\mathbf{Y}_\mathrm{emb}$ continues through the remainder of the decoder unchanged. Since MoFlow already predicts per-agent log-variance through its native uncertainty head, $\sigma$ is a first-class signal at this level, further motivating the tighter coupling.

\paragraph{Summary.}
The injection level is dictated by architecture, not by choice: MID and LED offer no intermediate future-agent embedding, so output-level is the natural and only option; MoFlow's MTR decoder structure creates an intermediate layer where the graph signal can synergize with the decoder's own agent-attention, making embedding-level injection both natural and strictly more expressive.

\vspace{4pt}

\noindent\textbf{Zero initialization.}
In all three hosts, SRA's output projection layer is initialized to zero weights and zero bias. This guarantees that at the start of fine-tuning, the module is a strict no-op: the host model's learned dynamics are preserved and the graph contribution is phased in gradually.

\noindent\textbf{First-step graph skip.}
At the very first denoising step there is no prior estimate $\hat{y}^{\tau}$ available from which to build interaction graph geometry. The graph is skipped (\texttt{skip\_graph=True}) for this step, and the host denoiser runs without relational correction. From the second step onward, the previous step's $\hat{y}^{\tau}$ provides graph geometry.

\noindent\textbf{Normalization for graph edges.}
Graph edge features are computed in absolute court/field coordinates (meters), not in the normalized training space. For MoFlow, this requires inverting the min-max normalization before passing $\hat{y}^{\tau}$ to the graph module. For MID and LED, the scale factor ($\text{TRAJ\_SCALE}=5$) and last observed position are added back before the graph receives the trajectory.

\subsection{From refined embedding to per-timestep residual: the decoder}
\label{sec:supp_decoder}

The aggregation step in Section~\ref{sec:unc_msg} produces a refined per-agent embedding $h_i^{\tau,(L)}\!\in\!\mathbb{R}^{D}$ for each agent (and each predicted mode, in multi-modal hosts). This embedding has no temporal axis: the per-timestep dynamics of the relational interaction were already absorbed into the edge features through attention pooling (Eq.~\ref{eq:edge_temporal}), and the message-passing aggregator operates strictly at the agent level. The decoder $\phi^{\text{dec}}_\theta$ is the component that re-introduces time, mapping the per-agent vector back to a $T_f\!\times\!2$ trajectory residual $\delta_i^{\tau}$.

Two design steps make this work cleanly across hosts.

\paragraph{Embedding-level gated residual.}
Before any host-specific decoding, the graph module emits a residual on the embedding itself, not on the trajectory. Let $h_i^{\tau,(0)}$ be the original (host-provided) per-agent embedding fed into the GNN, and $h_i^{\tau,(L)}$ the embedding after $L$ message-passing layers. We compute the scalar sigmoid gate $g_i\!\in\![0,1]$ of Eq.~\ref{eq:gated_residual} from the host feature and the pooled graph message $\bar m_i$, and form the refined embedding
\begin{equation}
\tilde h_i^{\tau} = h_i^{\tau,(0)} + g_i\, W_\text{out}\,h_i^{\tau,(L)},
\quad
g_i = \mathrm{sigmoid}\!\left(W_g\,[\,h_i^{\tau,(0)};\,\bar m_i\,]\right),
\label{eq:emb_gate}
\end{equation}
where $W_g\!\in\!\mathbb{R}^{1\times 2D}$ and $W_\text{out}\!\in\!\mathbb{R}^{D\times D}$. Because $W_\text{out}$ is zero-initialized, $\tilde h_i^{\tau}\!\equiv\!h_i^{\tau,(0)}$ at the start of training, and the graph contribution is phased in gradually.

\paragraph{Host-specific time decoder.}
The refined embedding $\tilde h_i^{\tau}$ is then mapped to the per-timestep residual $\delta_i^{\tau}\!\in\!\mathbb{R}^{T_f\times 2}$. The decoder differs by host:

\begin{itemize}
\item \textbf{LED and MID} (output-level injection) use a small dedicated MLP that expands the agent-level embedding into a flattened trajectory residual, then reshapes:
\begin{equation}
\delta_i^{\tau} = \mathrm{reshape}_{T_f\times 2}\!\Big(\phi^{\text{dec}}_\theta(\tilde h_i^{\tau})\Big),
\quad
\phi^{\text{dec}}_\theta:\,\mathbb{R}^{D}\!\rightarrow\!\mathbb{R}^{T_f\cdot 2},
\end{equation}
where $\phi^{\text{dec}}_\theta$ is a two-layer MLP: $\mathrm{Linear}(D\!\to\!D)\!\to\!\mathrm{ReLU}\!\to\!\mathrm{Linear}(D\!\to\!T_f\!\cdot\!2)$. The final linear layer is zero-initialized so the residual starts at zero.

\item \textbf{MoFlow} (embedding-level injection) does \emph{not} have a separate trajectory decoder inside the graph module: $\tilde h_i^{\tau}$ \emph{replaces} the host's intermediate embedding and is fed back through the remaining MTR decoder layers (A-attention, context-fusion MLP, motion decoder, regression head). The host's own regression head, which already maps a $D$-dim node embedding to $T_f\!\cdot\!2$ trajectory coordinates, plays the role of $\phi^{\text{dec}}_\theta$. No additional decoder is added.
\end{itemize}

\paragraph{Why no temporal transformer.}
A more elaborate alternative would have been a temporal transformer over node features after the GNN, mirroring the edge-side $\Psi^{\text{edge}}$. We tried this but found no measurable improvement: the relational temporal structure is already captured inside the per-edge feature $e_{ij}^{\tau}$ via attention pooling, and a second temporal transformer on per-agent features mostly added latency without changing accuracy. We accordingly use the simpler decoders above. This is also consistent with the host's existing architecture: MID's transformer denoiser, LED's frozen core, and MoFlow's MTR decoder all already produce per-timestep outputs from per-agent embeddings, so re-introducing a temporal mixer inside SRA would be redundant.

\raggedbottom
\subsection{Training and Inference Algorithms}
\label{sec:supp_algs}

The following algorithms describe how SRA integrates into the training and inference loops of each host denoiser. We use $\hat{y}^{\tau}$ for the current clean-trajectory estimate, $\sigma$ for per-agent uncertainty (log-variance), $g$ for the learnable gate, and $\delta$ for the graph residual. Notation follows Section~3 of the main paper.

\subsubsection*{MID (Diffusion DDPM)}

\begin{algorithm}[H]
\caption{SRA + MID: Training (one mini-batch iteration)}
\label{alg:mid_train}
\begin{algorithmic}[1]
\REQUIRE Past trajectories $x$, ground-truth future $y_0$; host transformer $\theta$, graph module $G$.
\STATE Compute 6-channel past features $p = [x_\mathrm{abs},\; x_\mathrm{rel},\; x_\mathrm{vel}]$
\STATE Encode past context: $c = \mathrm{Encoder}_\theta(p,\; \mathrm{mask})$
\STATE Sample diffusion level $\tau \sim \mathcal{U}\{1,\ldots,T\}$, noise $\varepsilon \sim \mathcal{N}(0,I)$
\STATE Forward diffuse: $y_\tau = \sqrt{\bar\alpha_\tau}\cdot y_0 + \sqrt{1-\bar\alpha_\tau}\cdot\varepsilon$
\STATE \textbf{Pass 1} \COMMENT{no gradient; build graph geometry from GT future (teacher forcing)}
\STATE \quad $\hat{y}^{(1)}_\mathrm{abs} \leftarrow y_0 \cdot \text{TRAJ\_SCALE} + \text{last\_obs}$ \COMMENT{absolute court coords}
\STATE \quad $\varepsilon^{(1)},\; s \leftarrow \mathrm{GraphDenoiserNet}_\theta(y_\tau,\,\beta_\tau,\,c;\; \hat{y}^{(1)}_\mathrm{abs},\;\sigma{=}\text{None})$
\STATE \quad $\sigma \leftarrow \exp(s/2)$ \COMMENT{per-agent uncertainty from log-variance head}
\STATE \textbf{Pass 2} \COMMENT{with gradient; use Pass 1 estimate for graph geometry}
\STATE \quad $\hat{y}^{(2)} \leftarrow \bigl(y_\tau - \sqrt{1-\bar\alpha_\tau}\cdot\varepsilon^{(1)}\bigr)/\sqrt{\bar\alpha_\tau}$ \COMMENT{recover $\hat y_0$ from Pass 1}
\STATE \quad $\hat{y}^{(2)}_\mathrm{abs} \leftarrow \hat{y}^{(2)} \cdot \text{TRAJ\_SCALE} + \text{last\_obs}$
\STATE \quad $\varepsilon^{(2)},\; s \leftarrow \mathrm{GraphDenoiserNet}_\theta(y_\tau,\,\beta_\tau,\,c;\; \hat{y}^{(2)}_\mathrm{abs},\;\sigma)$
\STATE \hspace{6pt} \textit{(Inside net: transformer encodes $y_\tau$; graph $G$ maps $\hat{y}^{(2)}_\mathrm{abs}$ to $\delta$;} \par
         \hspace{18pt} \textit{output-level: $\varepsilon^{(2)} = \varepsilon^{(2)}_\mathrm{base} + g_i\,\delta$)}
\STATE $\mathcal{L}_\mathrm{diff} = \lVert\varepsilon^{(2)} - \varepsilon\rVert^2$ \COMMENT{host MSE loss}
\STATE $\mathcal{L}_\sigma = \frac{1}{2}\bigl(s + \lVert\varepsilon - \varepsilon^{(2)}_\mathrm{base}\rVert^2 e^{-s}\bigr)$ \COMMENT{heteroscedastic NLL}
\STATE $\mathcal{L} = \mathcal{L}_\mathrm{diff} + \lambda_\sigma\,\mathcal{L}_\sigma$; \quad backpropagate, update $\theta$ and $G$
\end{algorithmic}
\end{algorithm}

\begin{algorithm}[H]
\caption{SRA + MID: Inference (one scene, $K$ samples)}
\label{alg:mid_infer}
\begin{algorithmic}[1]
\REQUIRE Past trajectories $x$; $K$ samples, 10 DDIM steps (stride 10 over $T{=}100$).
\STATE Compute $p$, encode $c = \mathrm{Encoder}(p,\mathrm{mask})$
\STATE Initialize $K$ noisy futures: $y_T \sim \mathcal{N}(0,I)^{K\times A\times T_f\times 2}$; $\hat{y}^\mathrm{prev}_\mathrm{abs} \leftarrow \texttt{None}$
\FOR{$\tau = T,\; T{-}10,\; \ldots,\; 10$}
  \STATE \textbf{Pass 1} (no grad):
  \STATE \quad $\varepsilon^{(1)},\; s \leftarrow \mathrm{GraphDenoiserNet}(y_\tau,\beta_\tau,c;\;\hat{y}^\mathrm{prev}_\mathrm{abs},\;\sigma{=}\text{None},\;\texttt{skip}=(\hat{y}^\mathrm{prev}_\mathrm{abs}{=}\texttt{None}))$
  \STATE \quad $\sigma \leftarrow \exp(s/2)$
  \STATE \textbf{Pass 2} (no grad):
  \STATE \quad $\hat{y}^{(2)} \leftarrow (y_\tau - \sqrt{1-\bar\alpha_\tau}\cdot\varepsilon^{(1)})/\sqrt{\bar\alpha_\tau}$
  \STATE \quad $\hat{y}^{(2)}_\mathrm{abs} \leftarrow \hat{y}^{(2)}\cdot\text{TRAJ\_SCALE} + \text{last\_obs}$
  \STATE \quad $\varepsilon^{(2)},\; \_\; \leftarrow \mathrm{GraphDenoiserNet}(y_\tau,\beta_\tau,c;\;\hat{y}^{(2)}_\mathrm{abs},\;\sigma)$
  \STATE \hspace{6pt} \textit{(output-level injection: $\varepsilon^{(2)} = \varepsilon^{(2)}_\mathrm{base} + g_i\,\delta$)}
  \STATE $\hat{y}^{(2)}_0 \leftarrow (y_\tau - \sqrt{1-\bar\alpha_\tau}\cdot\varepsilon^{(2)})/\sqrt{\bar\alpha_\tau}$
  \STATE $\hat{y}^\mathrm{prev}_\mathrm{abs} \leftarrow \hat{y}^{(2)}_0\cdot\text{TRAJ\_SCALE} + \text{last\_obs}$ \COMMENT{update for next step}
  \STATE DDIM step: $y_{\tau-10} = \sqrt{\bar\alpha_{\tau-10}}\cdot\hat{y}^{(2)}_0 + \sqrt{1-\bar\alpha_{\tau-10}}\cdot\varepsilon^{(2)}$
\ENDFOR
\RETURN Unnormalize $\hat{y}^{(2)}_0$ to absolute court coordinates; return $K$ predicted futures
\end{algorithmic}
\end{algorithm}

\subsubsection*{LED (Leapfrog DDPM)}

\begin{algorithm}[H]
\caption{SRA + LED: Training (one mini-batch iteration)}
\label{alg:led_train}
\begin{algorithmic}[1]
\REQUIRE Past trajectories $x$, GT future $y_0$; \textbf{frozen} core denoiser $\Phi$, leapfrog initializer $\psi$, graph module $G$.
\STATE Compute 6-channel past $p$, social mask
\STATE Run initializer: $(\hat{s}_\mathrm{pred},\; \mu,\; \log\sigma^2) \leftarrow \mathrm{LEDInit}_\psi(p,\mathrm{mask})$
\STATE Scale samples: $\hat{s}_\mathrm{pred} \leftarrow e^{\log\sigma^2/2}\cdot\hat{s}_\mathrm{pred}\;/\;\mathrm{std}(\hat{s}_\mathrm{pred})$
\STATE Initialize $K{=}20$ modes: $\mathrm{loc} = \hat{s}_\mathrm{pred} + \mu\,[\mathrm{None}]$; $\sigma = \log\sigma^2$
\FOR{each half $h \in \{\mathrm{loc}_{:,:10},\;\mathrm{loc}_{:,10:}\}$}
  \STATE $y_\mathrm{cur} \leftarrow h$
  \FOR{$\tau = 4, 3, 2, 1, 0$}
    \STATE $\varepsilon_\theta \leftarrow \Phi(y_\mathrm{cur},\,\beta_\tau,\,p,\,\mathrm{mask})$ \COMMENT{FROZEN core denoiser}
    \STATE $\hat{y}^\tau \leftarrow (y_\mathrm{cur} - \sqrt{1-\bar\alpha_\tau}\cdot\varepsilon_\theta)/\sqrt{\bar\alpha_\tau}$
    \STATE $\delta \leftarrow G(\hat{y}^\tau,\;x,\;\tau,\;\sigma)$
    \STATE $\varepsilon_\theta' \leftarrow \varepsilon_\theta - \frac{\sqrt{\bar\alpha_\tau}}{\sqrt{1-\bar\alpha_\tau}}\cdot\delta$ \COMMENT{output-level; delta in $\hat{y}_0$ space}
    \STATE $y_\mathrm{cur} \leftarrow \frac{1}{\sqrt{\alpha_\tau}}\bigl(y_\mathrm{cur} - \frac{1-\alpha_\tau}{\sqrt{1-\bar\alpha_\tau}}\varepsilon_\theta'\bigr) + \sqrt{\beta_\tau}\cdot\xi\cdot\epsilon_0$ \COMMENT{$\xi \ll 1$}
  \ENDFOR
\ENDFOR
\STATE $\hat{y}_\mathrm{pred} = \mathrm{concat}(y_\mathrm{cur,0},\; y_\mathrm{cur,1})$
\STATE $\mathcal{L}_\mathrm{init} = \text{trajectory regression loss}(\hat{y}_\mathrm{pred},\; y_0)$
\STATE $\mathcal{L}_\sigma = \frac{1}{2}(\log\sigma^2 + \lVert y_0 - \mu\rVert^2 e^{-\log\sigma^2})$
\STATE $\mathcal{L} = \mathcal{L}_\mathrm{init} + \lambda_\sigma\,\mathcal{L}_\sigma$; \quad backpropagate through $\psi$ and $G$ only (core denoiser $\Phi$ frozen)
\end{algorithmic}
\end{algorithm}

\begin{algorithm}[H]
\caption{SRA + LED: Inference (one scene)}
\label{alg:led_infer}
\begin{algorithmic}[1]
\REQUIRE Past trajectories $x$; $K{=}20$ modes, 5 leapfrog steps.
\STATE Compute $p$, mask
\STATE $(\hat{s}_\mathrm{pred},\; \mu,\; \log\sigma^2) \leftarrow \mathrm{LEDInit}(p,\mathrm{mask})$; scale, form $\mathrm{loc}$, $\sigma$
\FOR{each half $h \in \{\mathrm{loc}_{:,:10},\;\mathrm{loc}_{:,10:}\}$}
  \STATE $y_\mathrm{cur} \leftarrow h$
  \FOR{$\tau = 4, 3, 2, 1, 0$}
    \STATE $\varepsilon_\theta \leftarrow \Phi(y_\mathrm{cur},\,\beta_\tau,\,p,\,\mathrm{mask})$
    \STATE $\hat{y}^\tau \leftarrow (y_\mathrm{cur} - \sqrt{1-\bar\alpha_\tau}\cdot\varepsilon_\theta)/\sqrt{\bar\alpha_\tau}$
    \STATE $\delta \leftarrow G(\hat{y}^\tau,\; x,\; \tau,\; \sigma)$
    \STATE $\varepsilon_\theta' \leftarrow \varepsilon_\theta - \frac{\sqrt{\bar\alpha_\tau}}{\sqrt{1-\bar\alpha_\tau}}\cdot\delta$ \COMMENT{output-level injection}
    \STATE $y_\mathrm{cur} \leftarrow \frac{1}{\sqrt{\alpha_\tau}}\!\left(y_\mathrm{cur} - \frac{1-\alpha_\tau}{\sqrt{1-\bar\alpha_\tau}}\varepsilon_\theta'\right) + \sqrt{\beta_\tau}\cdot\xi\cdot\epsilon_0$
  \ENDFOR
\ENDFOR
\RETURN $\mathrm{concat}(y_\mathrm{cur,0},\; y_\mathrm{cur,1})\cdot\text{TRAJ\_SCALE} + \text{last\_obs}$ \quad ($K{=}20$ absolute futures)
\end{algorithmic}
\end{algorithm}

\subsubsection*{MoFlow (Flow Matching)}

\begin{algorithm}[H]
\caption{SRA + MoFlow: Training (one mini-batch iteration)}
\label{alg:moflow_train}
\begin{algorithmic}[1]
\REQUIRE Past trajectories $x$, GT future $y_0$ (min-max normalized to $[-1,1]$);
         MTR encoder $E$, motion transformer $M$ (includes graph $G$ and logvar head).
\STATE Encode past context: $\mathbf{x}_\mathrm{data} = E(x)$
\STATE Sample time: $t \sim \mathrm{LogitNormal}(-0.5,\;1.5)$
\STATE Sample noise: $\varepsilon \sim \mathcal{N}(0,I)$
\STATE Linear interpolation: $y_t = (1-t)\cdot\varepsilon + t\cdot y_0$ \COMMENT{flow-matching trajectory}
\STATE \textbf{Pass 1} \COMMENT{no gradient; teacher-forcing with $y_0$}
\STATE \quad $y_{0,\mathrm{abs}}^\mathrm{GT} \leftarrow \mathrm{unnorm}(y_0) + \text{last\_obs}$ \COMMENT{absolute coords for graph edges}
\STATE \quad $(\hat{y}^{(1)},\;\_,\;\ell^{(1)}) \leftarrow M.\mathrm{forward}(y_t,\;t,\;\mathbf{x}_\mathrm{data};\; y_{0,\mathrm{abs}}^\mathrm{GT},\;\sigma{=}\mathrm{None})$
\STATE \hspace{6pt} \textit{(Inside $M$: $K$-attn $\to$ [\,\textbf{embedding-level}: $\mathbf{Y}_\mathrm{emb} \leftarrow G(\mathbf{Y}_\mathrm{emb},\; y_{0,\mathrm{abs}}^\mathrm{GT},\ldots)$\,] $\to$ $A$-attn $\to$ decoder $\to$ $\hat{y}^{(1)}$)}
\STATE \quad $\sigma \leftarrow \bigl(\ell^{(1)}\bigr)^{1/2}$ \COMMENT{per-agent uncertainty from Pass 1 logvar}
\STATE \textbf{Pass 2} \COMMENT{with gradient; use $\hat{y}^{(1)}$ for graph geometry}
\STATE \quad $\hat{y}^{(1)}_\mathrm{abs} \leftarrow \mathrm{unnorm}(\hat{y}^{(1)}) + \text{last\_obs}$
\STATE \quad $(v_\mathrm{pred},\;\hat{c},\;\ell^{(2)}) \leftarrow M.\mathrm{forward}(y_t,\;t,\;\mathbf{x}_\mathrm{data};\; \hat{y}^{(1)}_\mathrm{abs},\;\sigma)$
\STATE \hspace{6pt} \textit{(Inside $M$: $K$-attn $\to$ [\,\textbf{embedding-level}: $\mathbf{Y}_\mathrm{emb} \leftarrow G(\mathbf{Y}_\mathrm{emb},\; \hat{y}^{(1)}_\mathrm{abs},\;\sigma,\ldots)$\,] $\to$ $A$-attn $\to$ decoder $\to$ $v_\mathrm{pred}$)}
\STATE Target velocity: $v^* = (y_0 - y_t)/(1-t)$
\STATE $\mathcal{L}_\mathrm{reg} = \lVert v_\mathrm{pred} - v^*\rVert^2$ \COMMENT{flow-matching regression}
\STATE $\mathcal{L}_\mathrm{cls} = \text{mode classification loss}(\hat{c},\; y_0)$
\STATE $\mathcal{L}_\sigma = \text{NLL}(\ell^{(2)},\; y_0)$ \COMMENT{MoFlow native uncertainty loss}
\STATE $\mathcal{L} = \mathcal{L}_\mathrm{reg} + \mathcal{L}_\mathrm{cls} + 0.01\cdot\mathcal{L}_\sigma$; \quad backpropagate, update $E$ and $M$
\end{algorithmic}
\end{algorithm}

\begin{algorithm}[H]
\caption{SRA + MoFlow: Inference (one scene, $K$ samples)}
\label{alg:moflow_infer}
\begin{algorithmic}[1]
\REQUIRE Past trajectories $x$; $K{=}20$ samples, 10 Euler steps ($\mathrm{d}t{=}0.1$).
\STATE Encode past: $\mathbf{x}_\mathrm{data} = E(x)$
\STATE Initialize: $y_T \sim \mathcal{N}(0,I)^{B\times K\times A\times T_f\times 2}$ \COMMENT{tied noise across $K$ modes}; $\hat{y}^\mathrm{prev}_\mathrm{abs} \leftarrow \texttt{None}$
\FOR{$s = 0, 1, \ldots, 9$ \quad (current time $t_s = s/10$)}
  \STATE Apply input scaling: $y_{t_s}^{\mathrm{in}} \leftarrow y_{t_s} \cdot \mathrm{scale}(t_s)$
  \STATE \textbf{Pass 1} (no grad):
  \STATE \quad $(\hat{y}^{(1)},\;\_,\;\ell^{(1)}) \leftarrow M.\mathrm{forward}(y_{t_s}^{\mathrm{in}},\;t_s,\;\mathbf{x}_\mathrm{data};\;\hat{y}^\mathrm{prev}_\mathrm{abs},\;\sigma{=}\mathrm{None},\;\texttt{skip}{=}(\hat{y}^\mathrm{prev}_\mathrm{abs}{=}\texttt{None}))$
  \STATE \quad \textit{(embedding-level: $G$ applied between $K$-attn and $A$-attn inside $M$)}
  \STATE \quad $\sigma \leftarrow \bigl(\ell^{(1)}\bigr)^{1/2}$;\quad $\hat{y}^{(1)}_\mathrm{abs} \leftarrow \mathrm{unnorm}(\hat{y}^{(1)}) + \text{last\_obs}$
  \STATE \textbf{Pass 2} (no grad):
  \STATE \quad $(v_\mathrm{pred},\;\_,\;\_) \leftarrow M.\mathrm{forward}(y_{t_s}^{\mathrm{in}},\;t_s,\;\mathbf{x}_\mathrm{data};\;\hat{y}^{(1)}_\mathrm{abs},\;\sigma)$
  \STATE \quad \textit{(embedding-level with $\sigma$: $G$ refines $\mathbf{Y}_\mathrm{emb}$, then $A$-attn, decoder)}
  \STATE $\hat{y}_0 \leftarrow v_\mathrm{pred}$ \COMMENT{model directly predicts clean $y_0$}
  \STATE $\hat{y}^\mathrm{prev}_\mathrm{abs} \leftarrow \mathrm{unnorm}(\hat{y}_0) + \text{last\_obs}$ \COMMENT{update for next step}
  \STATE Euler update: $y_{t_{s+1}} \leftarrow y_{t_s} + v_\mathrm{pred}\cdot\mathrm{d}t$
\ENDFOR
\RETURN $\mathrm{unnorm}(\hat{y}_0) + \text{last\_obs}$ \quad ($K{=}20$ absolute future trajectories)
\end{algorithmic}
\end{algorithm}

\FloatBarrier
\clearpage
\flushbottom
\section{Additional Ablations}
\label{sec:supp_ablations}

\subsection{Number of neighbors $K_{\mathrm{nbr}}$}
\label{sec:supp_num_neighbors}

\FloatBarrier
\input{tables/ablation_num_neighbor}
Table~\ref{tab:num_neighbors} sweeps the top-$K_{\mathrm{nbr}}$ neighborhood size used in the sparse selection step (Sec.~\ref{sec:sparse}) over $\{3,5,10\}$ on the NBA dataset, for all three host denoisers.
The pattern is consistent: $K_{\mathrm{nbr}}{=}5$ is the best or near-best setting in every case, with $K_{\mathrm{nbr}}{=}3$ trailing slightly and $K_{\mathrm{nbr}}{=}10$ either flat (MID, MoFlow) or actively destabilizing (LED, where FDE jumps from 1.161 at $K_{\mathrm{nbr}}{=}5$ to 2.303 at $K_{\mathrm{nbr}}{=}10$).

This validates a key design intuition: the optimal neighborhood size is governed by \emph{interaction density}, not by the number of agents in the scene.
NBA has 11 agents, but each player effectively interacts with only a handful of teammates and nearby defenders, so $K_{\mathrm{nbr}}{=}5$ already covers the relevant peer set.
Larger $K_{\mathrm{nbr}}$ admits weakly-related agents whose noisy edge features dilute the aggregated message; only considering the core interactions results in more accurate predictions.
We therefore fix $K_{\mathrm{nbr}}{=}5$ as the default in all main experiments.

\FloatBarrier
\clearpage
\section{Joint Metrics}
\label{sec:supp_joint}

ADE and FDE take the minimum over modes independently for each agent, so they do not reflect whether the predicted agents are consistent with one another. Joint ADE and Joint FDE (JADE/JFDE)~\cite{weng2023joint} instead select the best mode jointly over all agents in the scene, and therefore improve only when the predicted agents are mutually consistent. We do not use collision rate: physical contact between players is a normal part of basketball, soccer, and football, so a collision counter would penalize predictions that are in fact correct.

\input{tables/joint_metrics}

SRA improves JADE and JFDE (Table~\ref{tab:joint_metrics}) for every host on NBA and Soccer, by up to 5.9\%, and for LED and MoFlow on Football, by up to 9.5\%. Because these metrics are computed on a single jointly selected mode, the improvement indicates that the corrections SRA applies are mutually consistent across agents rather than accurate only in isolation. MID on Football is the exception: its marginal ADE/FDE improve, but its joint metrics degrade (consistent over three evaluation runs), so on this pair SRA sharpens per-agent accuracy without improving single-sample joint coherence.

\FloatBarrier
\section{Comparison with Future-Aware Interaction Modules}
\label{sec:supp_future_aware}

Of the prior methods that use predicted futures~\cite{huang2023gameformer, fu2025moflowonestepflowmatching, jia2023towards}, MoFlow is already one of our hosts, and SRA improves it on all four benchmarks (Table~\ref{main_sports}).

\paragraph{Adapter-level comparison.}
GameFormer~\cite{huang2023gameformer} is the closest at the mechanism level, since its level-$k$ reasoning also conditions each agent on the predicted futures of the others. Because SRA is an adapter, we compare at the level of adapters: we implement level-$k$ reasoning as an adapter that receives the same tensor at the same insertion point, returns a gated residual on the same embedding, and follows the same training and evaluation protocol (NBA, minADE$_{20}$ / minFDE$_{20}$ at 4.0\,s), so that the comparison isolates \emph{how} the predicted future is used.

\input{tables/adapter_comparison}

As Table~\ref{tab:adapter_comparison} shows, level-$k$ reasoning is inconsistent as an adapter: it improves MID, is mixed on LED, and substantially degrades MoFlow. It reads the host's current estimate with no measure of how reliable that estimate is, the same condition that our cumulative ablation isolates (Table~\ref{tab:ablation_cumulative}), where ungated future-conditioned interaction helps some hosts and hurts others. SRA weights each message by the relative uncertainty of its two endpoints and improves all three hosts with about half the parameters.

\paragraph{Standalone comparison.}
\label{sec:supp_standalone}
The coarse-to-fine approach (C2F)~\cite{jia2023towards} refines its prediction along the temporal axis and has no module that consumes other agents' predicted futures, so there is no equivalent slot to swap. We therefore compare against C2F, and against standalone GameFormer, as complete predictors (Table~\ref{tab:standalone}). All entries use the same NBA split and the minADE$_{20}$ / minFDE$_{20}$ protocol.

\begin{table}[!ht]
\centering
\small
\caption{\textbf{Standalone future-aware predictors vs.\ hosts with SRA} on NBA (ADE / FDE). Percentages are the relative improvement of SRA over its host.}
\label{tab:standalone}
\begin{tabular}{@{}llcc@{}}
\toprule
Category & Method & ADE $\downarrow$ & FDE $\downarrow$ \\
\midrule
\multirow{2}{*}{Standalone future-aware} & GameFormer~\cite{huang2023gameformer} & 0.848 & 0.951 \\
 & C2F~\cite{jia2023towards} & 0.904 & 1.063 \\
\midrule
\multirow{6}{*}{Host + adapter} & MID~\cite{gu2022stochastic} & 1.037 & 1.472 \\
 & MID + SRA & 0.957 (+7.7\%) & 1.346 (+8.6\%) \\
 & LED~\cite{mao2023leapfrog} & 0.839 & 1.204 \\
 & LED + SRA & 0.778 (+7.3\%) & 1.161 (+3.6\%) \\
 & MoFlow~\cite{fu2025moflowonestepflowmatching} & 0.703 & 0.884 \\
 & MoFlow + SRA & \textbf{0.695} (+1.1\%) & \textbf{0.870} (+1.6\%) \\
\bottomrule
\end{tabular}
\end{table}

A standalone predictor is a fixed artifact: its accuracy is a property of that one architecture. An adapter has no such ceiling, since its gain is realized on whatever host it is attached to and rises with that host. Applied to our strongest host, SRA surpasses both standalone future-aware predictors in ADE and FDE.

\section{Robustness Across Scenes}
\label{sec:supp_robustness}

A natural concern is that strengthening contextual information could overfit specific scenes. SRA's added context is largely scene-independent: it takes no map, scene, or agent-identity input, and it is derived from the host's own estimate rather than retrieved from anything fitted to the training scenes. The module operates on the relative geometry between agent pairs, which makes it invariant to global translation and equivariant to agent permutation by construction, and it does not observe absolute positions in scene coordinates. What SRA learns is how relative future geometry and confidence between two agents determine the correction one applies to the other, a regularity that holds across scenes rather than within any one of them.

The empirical evidence is consistent with this. (1) SRA improves ADE on all 12 host--dataset combinations and FDE on 11 of 12 (Table~\ref{main_sports}), including SDD, whose top-down pedestrian motion differs from the sports benchmarks in domain, agent count, and capture modality; the same module transfers unchanged. (2) The gains are distributed rather than concentrated: SRA lowers per-scene ADE on 60--90\% of the test scenes for every host (Table~\ref{tab:winrate}), so the aggregate is not carried by a favorable subset. (3) Figure~\ref{fig:uncertainty_comparison} shows gains in every uncertainty bin, with the largest gains on the highest-$\sigma$ scenes.

\begin{table}[!ht]
\centering
\small
\caption{\textbf{Percentage of test scenes in which SRA lowers the host's ADE.}}
\label{tab:winrate}
\begin{tabular}{@{}lcccc@{}}
\toprule
Host & NBA & Soccer & Football & Average \\
\midrule
MID    & 74.7\% & 81.5\% & 71.0\% & 75.7\% \\
LED    & 87.3\% & 82.6\% & 89.7\% & 86.5\% \\
MoFlow & 66.6\% & 74.0\% & 59.7\% & 66.8\% \\
\bottomrule
\end{tabular}
\end{table}

\FloatBarrier
\section{Inference Latency}
\label{sec:supp_latency}

We measure the inference cost added by SRA on each host denoiser with an optimized implementation, on a single NVIDIA RTX 3090 GPU with batch size 16 and $K{=}20$ predicted modes on NBA.

\begin{table}[!ht]
\centering
\small
\caption{\textbf{Inference latency on NBA} ($K{=}20$, batch 16, single RTX 3090). Per-sample time is the per-batch time divided by the batch size; $\Delta$ is the overhead added by SRA.}
\label{tab:latency_b16}
\begin{tabular}{lcccc}
\toprule
\textbf{Method} & \textbf{Per-batch (ms)} & \textbf{Per-sample (ms)} & \textbf{$\Delta$ per-sample (ms)} & \textbf{$\Delta$ \%} \\
\midrule
LED               & 233.2 & 14.58 & --    & --      \\
LED + SRA         & 256.8 & 16.05 & +1.47 & +10.1\% \\
\midrule
MoFlow            & 86.0  & 5.37  & --    & --      \\
MoFlow + SRA      & 222.8 & 13.93 & +8.56 & +159.4\% \\
\midrule
MID               & 803.7 & 50.23 & --    & --      \\
MID + SRA         & 847.4 & 52.96 & +2.73 & +5.4\%  \\
\bottomrule
\end{tabular}
\end{table}

The computation added by SRA is structurally small. Rebuilding the graph is a two-term retrieval score over agent pairs (a dot product plus a geometric bias over four summary statistics, with no per-timestep encoding), and the expensive operations, relational encoding and message passing, run only on the $N\!\cdot\!K_{\mathrm{nbr}}$ retained edges rather than all $N(N{-}1)$ pairs, which is what sparse selection exists for. The module has 0.63\,M parameters, shared across all agents, edges, and reverse steps.

In absolute terms, SRA adds 1.5 to 8.6\,ms per sample. The largest relative figure, $+159\%$ on MoFlow, reflects how little the unmodified host costs rather than how much SRA costs: per-sample latency moves from 5.4\,ms to 13.9\,ms, which is still faster than either unmodified LED or MID. NBA is sampled at 5\,Hz, so a scene must be processed within 200\,ms for real-time operation, and the slowest configuration (MID + SRA) uses 53\,ms of that budget.

\FloatBarrier
\section{Limitations}
\label{sec:supp_limitations}

\paragraph{Training requirement.}
SRA is modular, not training-free. It can be integrated into an existing predictor without replacing or redesigning its generative core, but it requires a host-specific injection point (output-level for MID and LED, embedding-level for MoFlow; Sec.~\ref{sec:supp_injection}) and joint training or fine-tuning with the host. It cannot be attached to a frozen, pretrained host without any training.

\paragraph{Scope of compatible hosts.}
SRA assumes that the host (i) refines a continuous future estimate through an iterative process, (ii) exposes a prediction or intermediate representation into which a correction can be injected, and (iii) provides agent features from which an uncertainty head can be trained. We validate it on three hosts that span DDPM, leapfrog-DDPM, and flow matching; regression- and classification-based predictors without an iterative refinement chain are not covered, and extending SRA to them remains future work.

\paragraph{Inference latency overhead.}
SRA rebuilds a future-conditioned interaction graph at every reverse step, on top of the host's own computation. With the optimized implementation the overhead is 1.5--8.6\,ms per scene (Table~\ref{tab:latency_b16}), well inside the 200\,ms budget of 5\,Hz sports data, but the relative overhead is large for very fast hosts such as MoFlow ($+159\%$), and the all-pairs retrieval score grows as $\mathcal{O}(N^2)$ in the number of agents, which may matter for large crowds.

\paragraph{Dependence on the host denoiser.}
SRA constructs its graph from the host denoiser's intermediate future estimate, so the quality of the graph depends on the quality of that estimate.
If the host moves toward an incorrect mode at an early generation step, or mispredicts an interaction-critical agent, the adapter can build edges from an inaccurate future geometry.
Zero-initialized residuals, gated corrections, sparse neighborhoods, uncertainty weighting, and iterative graph recomputation reduce the impact of such errors but do not eliminate them.
The method does not provide a formal stability guarantee for the coupled host and adapter dynamics.

\paragraph{Metric coverage.}
We report marginal (ADE/FDE) and joint (JADE/JFDE) displacement metrics. Joint metrics capture whether the predicted agents are mutually consistent, but they still do not measure map violation, social-rule violation, physical feasibility, or downstream planning cost, and we do not report collision rate because physical contact is a normal part of the sports we evaluate. SRA improves the joint metrics in 8 of 9 sports host--dataset pairs; for MID on Football, joint metrics degrade although marginal metrics improve.

\paragraph{Dataset and scene scope.}
The experimental evaluation is concentrated on SDD and sports benchmarks.
These datasets primarily contain scenes with a single dominant agent class, such as pedestrians or sports players, and the relational encoder uses a shared agent-centric representation.
We do not evaluate mixed-class scenes with heterogeneous dynamics, physical footprints, right-of-way rules, or map-dependent interactions.
We also do not test robustness to perception noise, occlusion, identity switches, asynchronous sensing, or large-scale crowds.

\paragraph{Broader impact.}
More coherent multi-agent forecasts can make planners for autonomous vehicles and robots that share space with people safer and less conservative, and can support sports analytics. Conversely, forecasts that look coherent may be over-trusted: a downstream system should not treat SRA's predictions or its uncertainty estimates as guarantees, particularly outside the domains evaluated here.

\clearpage
\section{More Qualitative Results}
\label{sec:more_qual}
Figures~\ref{fig:supp_more1} and~\ref{fig:supp_more2} compare predictions with and without SRA on additional NBA scenes. With SRA, the predicted agents are more coherent with one another, because each prediction is refined against the predicted futures of its neighbors.
Figure~\ref{fig:supp_more3} shows LED with SRA on complex NBA scenes: agents in the middle of a crowded scene are assigned high uncertainty, and SRA predicts them more accurately.
\input{figures_latex/figure_supp_qual_more1}
\clearpage
\input{figures_latex/figure_supp_qual_more2}
\clearpage
\input{figures_latex/figure_supp_qual_uncertainty_more}
\clearpage

%% file: tables/ablation_num_neighbor.tex
\begin{wraptable}{r}{0.5\linewidth}
\centering
\scriptsize
\setlength{\tabcolsep}{3pt}
\begin{tabular}{@{}llcc@{}}
\toprule
Model & $K_{\mathrm{nbr}}$ & ADE & FDE \\
\midrule
\multirow{3}{*}{MID + SRA}    & 3  & 0.960 & 1.346 \\
                              & 5  & 0.957 & 1.346 \\
                              & 10 & 0.968 & 1.356 \\
\midrule
\multirow{3}{*}{LED + SRA}    & 3  & 0.775 & 1.163 \\
                              & 5  & 0.778 & 1.161 \\
                              & 10 & 0.783 & 2.303 \\
\midrule
\multirow{3}{*}{MoFlow + SRA} & 3  & 0.703 & 0.893 \\
                              & 5  & 0.695 & 0.870 \\
                              & 10 & 0.704 & 0.878 \\
\bottomrule
\end{tabular}
\caption{Ablation on the number of selected neighbors $K_{\mathrm{nbr}}$ on NBA (ADE / FDE, best of 20). $K_{\mathrm{nbr}}{=}5$ is the default.}
\label{tab:num_neighbors}
\end{wraptable}

%% file: tables/joint_metrics.tex
\begin{table}[!ht]
\centering
\small
\caption{\textbf{Joint metrics} JADE / JFDE~\cite{weng2023joint} on the sports benchmarks, which select one mode jointly for all agents in the scene. Better of each host pair in \textbf{bold}.}
\label{tab:joint_metrics}
\setlength{\tabcolsep}{4pt}
\begin{tabular}{@{}l cc cc cc@{}}
\toprule
& \multicolumn{2}{c}{NBA} & \multicolumn{2}{c}{Soccer} & \multicolumn{2}{c}{Football} \\
\cmidrule(lr){2-3} \cmidrule(lr){4-5} \cmidrule(lr){6-7}
Method & JADE & JFDE & JADE & JFDE & JADE & JFDE \\
\midrule
MID          & 1.977 & 3.706 & 0.789 & 1.577 & \textbf{0.581} & \textbf{1.184} \\
MID + SRA    & \textbf{1.945} & \textbf{3.597} & \textbf{0.755} & \textbf{1.518} & 0.603 & 1.238 \\
\midrule
LED          & 1.609 & 2.851 & 0.601 & 1.121 & 0.550 & 1.071 \\
LED + SRA    & \textbf{1.580} & \textbf{2.842} & \textbf{0.575} & \textbf{1.055} & \textbf{0.501} & \textbf{0.969} \\
\midrule
MoFlow       & 1.476 & 2.708 & 0.654 & 1.067 & 0.509 & 0.991 \\
MoFlow + SRA & \textbf{1.454} & \textbf{2.675} & \textbf{0.636} & \textbf{1.037} & \textbf{0.503} & \textbf{0.976} \\
\bottomrule
\end{tabular}
\end{table}

%% file: tables/adapter_comparison.tex
\begin{table}[!ht]
\centering
\small
\caption{\textbf{Adapter-level comparison} on NBA (ADE / FDE). GameFormer's level-$k$ reasoning~\cite{huang2023gameformer} is placed in the same slot as SRA, with the same input, injection point, and training protocol.}
\label{tab:adapter_comparison}
\begin{tabular}{@{}l ccc@{}}
\toprule
Host & Baseline & + level-$k$~\cite{huang2023gameformer} & + SRA (ours) \\
\midrule
MID (diffusion)        & 1.037 / 1.472 & 0.984 / 1.394 & \textbf{0.957} / \textbf{1.346} \\
LED (leapfrog)         & 0.839 / 1.204 & 0.836 / 1.266 & \textbf{0.778} / \textbf{1.161} \\
MoFlow (flow matching) & 0.703 / 0.884 & 0.949 / 1.098 & \textbf{0.695} / \textbf{0.870} \\
\midrule
\textit{Module parameters} & --- & 1.30\,M & \textbf{0.63\,M} \\
\bottomrule
\end{tabular}
\end{table}

%% file: figures_latex/figure_supp_qual_more1.tex
\begin{figure}[H]
    \includegraphics[width=1.0\linewidth]{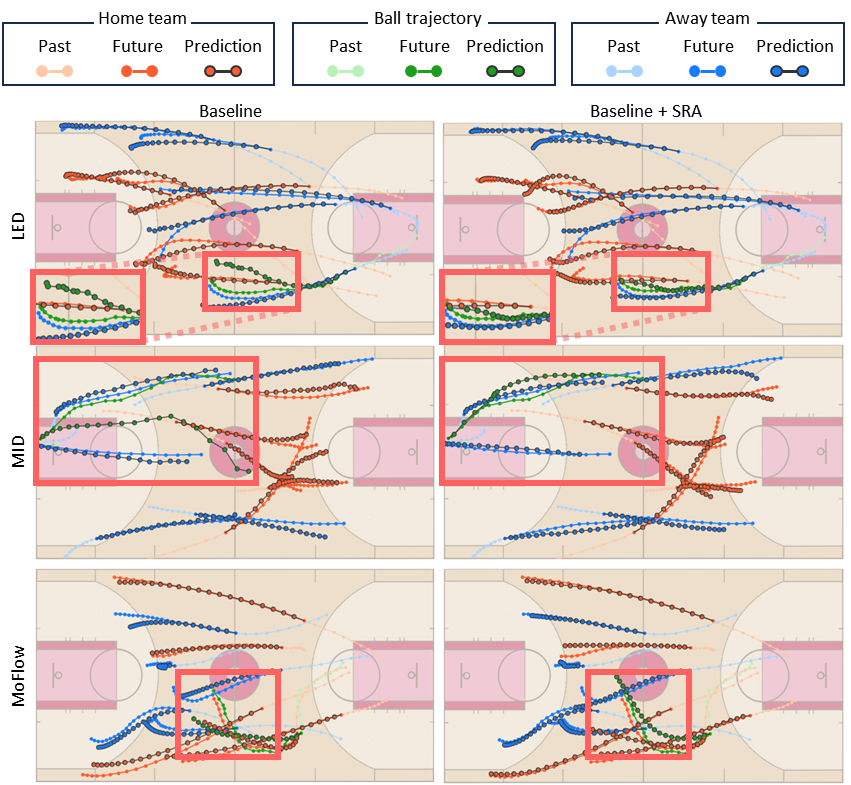} \vspace{-7pt}
    \caption{
    Additional qualitative comparisons on NBA, in the same format as Fig.~\ref{fig:qual_main}. Rows: hosts (LED, MID, MoFlow); left: host alone, right: host + SRA. Light, medium, and dark dots are the observed past, the ground-truth future, and the prediction (best of $K{=}20$ modes by ADE); red boxes mark where the predictions differ.
    }
    \label{fig:supp_more1}
    \vspace{-10pt}
\end{figure}

%% file: figures_latex/figure_supp_qual_more2.tex
\begin{figure}
    \centering
    \includegraphics[width=1.0\linewidth]{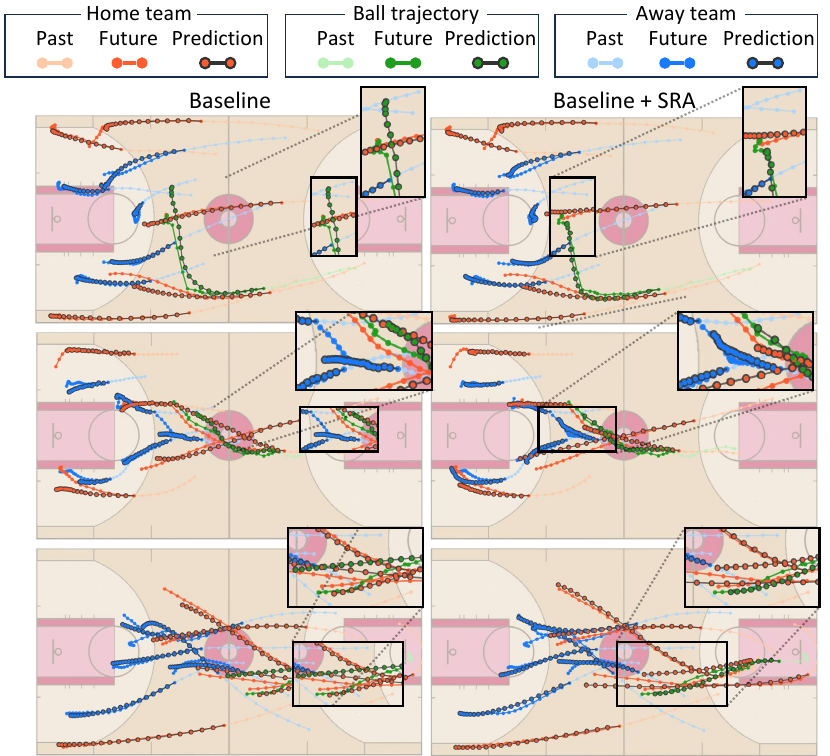} \vspace{-3pt}
    \caption{
    Additional qualitative comparisons on NBA. Rows (top to bottom): LED, MID, and MoFlow as hosts; left: host alone, right: host + SRA. Light, medium, and dark dots are the observed past, the ground-truth future, and the prediction (best of $K{=}20$ modes by ADE). Black boxes mark where the predictions differ and are enlarged in the insets.
    }
    \label{fig:supp_more2}
    \vspace{-10pt}
\end{figure}

%% file: figures_latex/figure_supp_qual_uncertainty_more.tex
\begin{figure}
    \includegraphics[width=1.0\linewidth]{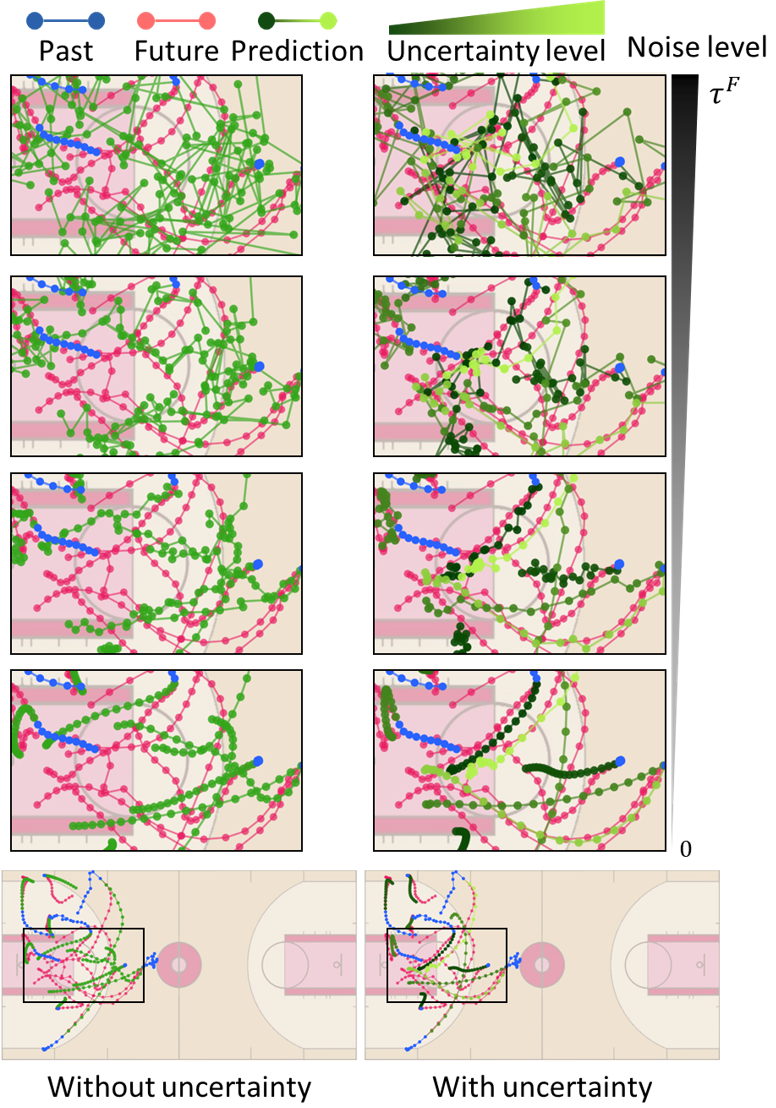}
    \vspace{-7pt}
    \caption{
    Additional denoising examples without (left) and with (right) uncertainty, LED on NBA, in the same format as Fig.~\ref{fig:qual_uncertainty}. Rows above the full scene (bottom) show the boxed region from high noise (top) to the final step; blue: past, pink: ground-truth future, green: prediction shaded by predicted uncertainty. Agents in the crowded center receive high uncertainty and are predicted more accurately with uncertainty-weighted message passing.
    }
    \label{fig:supp_more3}
    \vspace{-10pt}
\end{figure}

%% file: checklist.tex
\section*{NeurIPS Paper Checklist}

\begin{enumerate}

\item {\bf Claims}
    \item[] Question: Do the main claims made in the abstract and introduction accurately reflect the paper's contributions and scope?
    \item[] Answer: \answerYes{} 
    \item[] Justification: The paper states its main claims in the abstract and introduction, and supports them through the proposed method and empirical results.

\item {\bf Limitations}
    \item[] Question: Does the paper discuss the limitations of the work performed by the authors?
    \item[] Answer: \answerYes{} 
    \item[] Justification: Limitations are summarized in Sec.~\ref{sec:conclusion} and discussed in detail in Supp.~\ref{sec:supp_limitations} (training requirement, scope of compatible hosts, latency, dependence on the host estimate, metric coverage, and dataset scope).

\item {\bf Theory assumptions and proofs}
    \item[] Question: For each theoretical result, does the paper provide the full set of assumptions and a complete (and correct) proof?
    \item[] Answer: \answerNA{} 
    \item[] Justification: The paper does not include theoretical results.

    \item {\bf Experimental result reproducibility}
    \item[] Question: Does the paper fully disclose all the information needed to reproduce the main experimental results of the paper to the extent that it affects the main claims and/or conclusions of the paper (regardless of whether the code and data are provided or not)?
    \item[] Answer: \answerYes{} 
    \item[] Justification: The paper provides the method description, details, datasets, metrics, and evaluation protocol needed to reproduce the main results.

\item {\bf Open access to data and code}
    \item[] Question: Does the paper provide open access to the data and code, with sufficient instructions to faithfully reproduce the main experimental results, as described in supplemental material?
    \item[] Answer: \answerYes{} 
    \item[] Justification: Code is released at \url{https://github.com/Jaewoo97/SRA}; all datasets are public.

\item {\bf Experimental setting/details}
    \item[] Question: Does the paper specify all the training and test details (e.g., data splits, hyperparameters, how they were chosen, type of optimizer) necessary to understand the results?
    \item[] Answer: \answerYes{} 
    \item[] Justification: The paper specifies the experimental settings, including datasets, training details, evaluation metrics, and implementation choices.

\item {\bf Experiment statistical significance}
    \item[] Question: Does the paper report error bars suitably and correctly defined or other appropriate information about the statistical significance of the experiments?
    \item[] Answer: \answerNo{} 
    \item[] Justification: The paper does not report error bars.

\item {\bf Experiments compute resources}
    \item[] Question: For each experiment, does the paper provide sufficient information on the computer resources (type of compute workers, memory, time of execution) needed to reproduce the experiments?
    \item[] Answer: \answerYes{} 
    \item[] Justification: The paper reports the compute resources used for the experiments, including hardware and runtime information.
    
\item {\bf Code of ethics}
    \item[] Question: Does the research conducted in the paper conform, in every respect, with the NeurIPS Code of Ethics \url{https://neurips.cc/public/EthicsGuidelines}?
    \item[] Answer: \answerYes{} 
    \item[] Justification: The research follows the NeurIPS Code of Ethics and does not involve deceptive, harmful, or privacy-invasive data collection.

\item {\bf Broader impacts}
    \item[] Question: Does the paper discuss both potential positive societal impacts and negative societal impacts of the work performed?
    \item[] Answer: \answerYes{} 
    \item[] Justification: The broader-impact paragraph in Supp.~\ref{sec:supp_limitations} discusses potential positive impacts for safer trajectory forecasting as well as the risk of over-trusting coherent-looking forecasts in safety-critical systems.
    
\item {\bf Safeguards}
    \item[] Question: Does the paper describe safeguards that have been put in place for responsible release of data or models that have a high risk for misuse (e.g., pre-trained language models, image generators, or scraped datasets)?
    \item[] Answer: \answerNA{} 
    \item[] Justification: The paper does not release high-risk models or datasets that require special safeguards.

\item {\bf Licenses for existing assets}
    \item[] Question: Are the creators or original owners of assets (e.g., code, data, models), used in the paper, properly credited and are the license and terms of use explicitly mentioned and properly respected?
    \item[] Answer: \answerNo{} 
    \item[] Justification: The paper cites the existing datasets and codebases used in the experiments, but does not explicitly list all licenses and terms of use.

\item {\bf New assets}
    \item[] Question: Are new assets introduced in the paper well documented and is the documentation provided alongside the assets?
    \item[] Answer: \answerYes{} 
    \item[] Justification: The only new asset is our code, released at \url{https://github.com/Jaewoo97/SRA} with documentation for setup, training, and evaluation. The paper introduces no new datasets.

\item {\bf Crowdsourcing and research with human subjects}
    \item[] Question: For crowdsourcing experiments and research with human subjects, does the paper include the full text of instructions given to participants and screenshots, if applicable, as well as details about compensation (if any)? 
    \item[] Answer: \answerNA{} 
    \item[] Justification: The paper does not involve crowdsourcing or research with human subjects.

\item {\bf Institutional review board (IRB) approvals or equivalent for research with human subjects}
    \item[] Question: Does the paper describe potential risks incurred by study participants, whether such risks were disclosed to the subjects, and whether Institutional Review Board (IRB) approvals (or an equivalent approval/review based on the requirements of your country or institution) were obtained?
    \item[] Answer: \answerNA{} 
    \item[] Justification: The paper does not involve crowdsourcing or research with human subjects, so IRB approval is not applicable.

\item {\bf Declaration of LLM usage}
    \item[] Question: Does the paper describe the usage of LLMs if it is an important, original, or non-standard component of the core methods in this research? Note that if the LLM is used only for writing, editing, or formatting purposes and does \emph{not} impact the core methodology, scientific rigor, or originality of the research, declaration is not required.
    \item[] Answer: \answerNA{} 
    \item[] Justification: The core method does not use LLMs as an important, original, or non-standard component.

\end{enumerate}

%% file: main.bib
@String(AAAI = {AAAI})

@inproceedings{ngiam2021scene,
  title={{Scene Transformer}: A unified architecture for predicting multiple agent trajectories},
  author={Ngiam, Jiquan and Caine, Benjamin and Vasudevan, Vijay and Zhang, Zhengdong and Chiang, Hao-Tien Lewis and Ling, Jeffrey and Roelofs, Rebecca and Bewley, Alex and Liu, Chenxi and Venugopal, Ashish and others},
  booktitle={International Conference on Learning Representations},
  year={2022}
}

@inproceedings{liu2024laformer,
  title={{LAformer}: Trajectory prediction for autonomous driving with lane-aware scene constraints},
  author={Liu, Mengmeng and Cheng, Hao and Chen, Lin and Broszio, Hellward and Li, Jiangtao and Zhao, Runjiang and Sester, Monika and Yang, Michael Ying},
  booktitle={Proceedings of the IEEE/CVF Conference on Computer Vision and Pattern Recognition Workshops},
  pages={2039--2049},
  year={2024}
}

@inproceedings{gu2024producing,
  title={Producing and leveraging online map uncertainty in trajectory prediction},
  author={Gu, Xunjiang and Song, Guanyu and Gilitschenski, Igor and Pavone, Marco and Ivanovic, Boris},
  booktitle={Proceedings of the IEEE/CVF Conference on Computer Vision and Pattern Recognition},
  pages={14521--14530},
  year={2024}
}

@article{wu2023ar,
  title={{AR-Diffusion}: Auto-regressive diffusion model for text generation},
  author={Wu, Tong and Fan, Zhihao and Liu, Xiao and Zheng, Hai-Tao and Gong, Yeyun and Shen, Yelong and Jiao, Jian and Li, Juntao and Wei, Zhongyu and Guo, Jian and Duan, Nan and Chen, Weizhu},
  journal={Advances in Neural Information Processing Systems},
  volume={36},
  pages={39957--39974},
  year={2023}
}

@inproceedings{wewer25srm,
  title     = {Spatial Reasoning with Denoising Models},
  author    = {Wewer, Christopher and Pogodzinski, Bartlomiej and Schiele, Bernt and Lenssen, Jan Eric},
  booktitle = {International Conference on Machine Learning},
  year      = {2025},
}

@InProceedings{pmlr-v235-ruhe24a,
  title = 	 {Rolling Diffusion Models},
  author =       {Ruhe, David and Heek, Jonathan and Salimans, Tim and Hoogeboom, Emiel},
  booktitle = 	 {Proceedings of the 41st International Conference on Machine Learning},
  pages = 	 {42818--42835},
  year = 	 {2024},
  editor = 	 {Salakhutdinov, Ruslan and Kolter, Zico and Heller, Katherine and Weller, Adrian and Oliver, Nuria and Scarlett, Jonathan and Berkenkamp, Felix},
  volume = 	 {235},
  series = 	 {Proceedings of Machine Learning Research},
  month = 	 {21--27 Jul},
  publisher =    {PMLR},
  url = 	 {https://proceedings.mlr.press/v235/ruhe24a.html}
}

@inproceedings{zhang2024tedi,
  title={{TEDi}: Temporally-entangled diffusion for long-term motion synthesis},
  author={Zhang, Zihan and Liu, Richard and Hanocka, Rana and Aberman, Kfir},
  booktitle={ACM SIGGRAPH 2024 Conference Papers},
  pages={1--11},
  year={2024}
}

@article{chen2024diffusion,
  title={Diffusion forcing: Next-token prediction meets full-sequence diffusion},
  author={Chen, Boyuan and Mart{\'\i} Mons{\'o}, Diego and Du, Yilun and Simchowitz, Max and Tedrake, Russ and Sitzmann, Vincent},
  journal={Advances in Neural Information Processing Systems},
  volume={37},
  pages={24081--24125},
  year={2024}
}

@INPROCEEDINGS{luo2021safety,
  title={Safety-oriented pedestrian occupancy forecasting},
  author={Luo, Katie and Casas, Sergio and Liao, Renjie and Yan, Xinchen and Xiong, Yuwen and Zeng, Wenyuan and Urtasun, Raquel},
  booktitle={2021 IEEE/RSJ International Conference on Intelligent Robots and Systems (IROS)},
  pages={1015--1022},
  year={2021},
  organization={IEEE}
}

@inproceedings{nayakanti2022way,
  title={{Wayformer}: Motion forecasting via simple \& efficient attention networks},
  author={Nayakanti, Nigamaa and Al-Rfou, Rami and Zhou, Aurick and Goel, Kratarth and Refaat, Khaled S and Sapp, Benjamin},
  booktitle={IEEE International Conference on Robotics and Automation (ICRA)},
  pages={2980--2987},
  year={2023}
}

@article{phong2023truly,
  title={What truly matters in trajectory prediction for autonomous driving?},
  author={Phong, Tran and Wu, Haoran and Yu, Cunjun and Cai, Panpan and Zheng, Sifa and Hsu, David},
  journal={Advances in Neural Information Processing Systems},
  volume={36},
  pages={71327--71339},
  year={2023}
}

@inproceedings{lin2024giga,
  title={{GigaTraj}: Predicting long-term trajectories of hundreds of pedestrians in gigapixel complex scenes},
  author={Lin, Haozhe and Wei, Chunyu and He, Li and Guo, Yuchen and Zhao, Yunqi and Li, Shanglong and Fang, Lu},
  booktitle={Proceedings of the IEEE/CVF Conference on Computer Vision and Pattern Recognition},
  pages={19331--19340},
  year={2024}
}

@inproceedings{liao2024cdstraj,
  title={{CDSTraj}: Characterized diffusion and spatial-temporal interaction network for trajectory prediction in autonomous driving},
  author={Liao, Haicheng and Li, Xuelin and Li, Yongkang and Kong, Hanlin and Wang, Chengyue and Wang, Bonan and Guan, Yanchen and Tam, K and Li, Zhenning},
  booktitle={Proceedings of the Thirty-Third International Joint Conference on Artificial Intelligence, IJCAI-24},
  pages={7331--7339},
  year={2024}
}

@InProceedings{Jeong_2025_CVPR,
  title={Multi-modal Knowledge Distillation-based Human Trajectory Forecasting},
  author={Jeong, Jaewoo and Lee, Seohee and Park, Daehee and Lee, Giwon and Yoon, Kuk-Jin},
  booktitle={Proceedings of the Computer Vision and Pattern Recognition Conference},
  pages={24222--24233},
  year={2025}
}

@inproceedings{zhou2023qcnet,
  title={Query-centric trajectory prediction},
  author={Zhou, Zikang and Wang, Jianping and Li, Yung-Hui and Huang, Yu-Kai},
  booktitle={Proceedings of the IEEE/CVF Conference on Computer Vision and Pattern Recognition},
  pages={17863--17873},
  year={2023}
}

@InProceedings{Wong2024SoCir,
  title={{SocialCircle}: Learning the angle-based social interaction representation for pedestrian trajectory prediction},
  author={Wong, Conghao and Xia, Beihao and Zou, Ziqian and Wang, Yulong and You, Xinge},
  booktitle={Proceedings of the IEEE/CVF Conference on Computer Vision and Pattern Recognition},
  pages={19005--19015},
  year={2024}
}

@inproceedings{mao2023leapfrog,
  title={Leapfrog diffusion model for stochastic trajectory prediction},
  author={Mao, Weibo and Xu, Chenxin and Zhu, Qi and Chen, Siheng and Wang, Yanfeng},
  booktitle={Proceedings of the IEEE/CVF Conference on Computer Vision and Pattern Recognition},
  pages={5517--5526},
  year={2023}
}

@article{salzmann2023robots,
  title={Robots that can see: Leveraging human pose for trajectory prediction},
  author={Salzmann, Tim and Chiang, Hao-Tien Lewis and Ryll, Markus and Sadigh, Dorsa and Parada, Carolina and Bewley, Alex},
  journal={IEEE Robotics and Automation Letters},
  volume={8},
  number={11},
  pages={7090--7097},
  year={2023},
  publisher={IEEE}
}

@InProceedings{Xu2022RMBTP,
  title={Remember intentions: Retrospective-memory-based trajectory prediction},
  author={Xu, Chenxin and Mao, Weibo and Zhang, Wenjun and Chen, Siheng},
  booktitle={Proceedings of the IEEE/CVF Conference on Computer Vision and Pattern Recognition},
  pages={6488--6497},
  year={2022}
}

@article{Chib2024PTD,
  title={Pedestrian trajectory prediction with missing data: Datasets, imputation, and benchmarking},
  author={Chib, Pranav Singh and Singh, Pravendra},
  journal={Advances in Neural Information Processing Systems},
  volume={37},
  pages={124530--124546},
  year={2024}
}

@InProceedings{Pang2021TPL,
  title={Trajectory prediction with latent belief energy-based model},
  author={Pang, Bo and Zhao, Tianyang and Xie, Xu and Wu, Ying Nian},
  booktitle={Proceedings of the IEEE/CVF Conference on Computer Vision and Pattern Recognition},
  pages={11814--11824},
  year={2021}
}

@InProceedings{Gupta2018SGAN,
  title={Social {GAN}: Socially acceptable trajectories with generative adversarial networks},
  author={Gupta, Agrim and Johnson, Justin and Fei-Fei, Li and Savarese, Silvio and Alahi, Alexandre},
  booktitle={Proceedings of the IEEE Conference on Computer Vision and Pattern Recognition},
  pages={2255--2264},
  year={2018}
}

@InProceedings{Fang2020TPnet,
  title={{TPNet}: Trajectory proposal network for motion prediction},
  author={Fang, Liangji and Jiang, Qinhong and Shi, Jianping and Zhou, Bolei},
  booktitle={Proceedings of the IEEE/CVF Conference on Computer Vision and Pattern Recognition},
  pages={6797--6806},
  year={2020}
}

@InProceedings{Liang2020FoP,
  title={The garden of forking paths: Towards multi-future trajectory prediction},
  author={Liang, Junwei and Jiang, Lu and Murphy, Kevin and Yu, Ting and Hauptmann, Alexander},
  booktitle={Proceedings of the IEEE/CVF Conference on Computer Vision and Pattern Recognition},
  pages={10508--10518},
  year={2020}
}

@InProceedings{Sun2020RLN,
  title={Reciprocal learning networks for human trajectory prediction},
  author={Sun, Hao and Zhao, Zhiqun and He, Zhihai},
  booktitle={Proceedings of the IEEE/CVF Conference on Computer Vision and Pattern Recognition},
  pages={7416--7425},
  year={2020}
}

@inproceedings{alahi2016social,
  title={Social {LSTM}: Human trajectory prediction in crowded spaces},
  author={Alahi, Alexandre and Goel, Kratarth and Ramanathan, Vignesh and Robicquet, Alexandre and Fei-Fei, Li and Savarese, Silvio},
  booktitle={Proceedings of the IEEE Conference on Computer Vision and Pattern Recognition},
  pages={961--971},
  year={2016}
}

@inproceedings{hu2020collaborative,
  title={Collaborative motion prediction via neural motion message passing},
  author={Hu, Yue and Chen, Siheng and Zhang, Ya and Gu, Xiao},
  booktitle={Proceedings of the IEEE/CVF Conference on Computer Vision and Pattern Recognition},
  pages={6319--6328},
  year={2020}
}

@article{tang2021collaborative,
  title={Collaborative uncertainty in multi-agent trajectory forecasting},
  author={Tang, Bohan and Zhong, Yiqi and Neumann, Ulrich and Wang, Gang and Chen, Siheng and Zhang, Ya},
  journal={Advances in Neural Information Processing Systems},
  volume={34},
  pages={6328--6340},
  year={2021}
}

@inproceedings{xu2022groupnet,
  title={{GroupNet}: Multiscale hypergraph neural networks for trajectory prediction with relational reasoning},
  author={Xu, Chenxin and Li, Maosen and Ni, Zhenyang and Zhang, Ya and Chen, Siheng},
  booktitle={Proceedings of the IEEE/CVF Conference on Computer Vision and Pattern Recognition},
  pages={6498--6507},
  year={2022}
}

@inproceedings{lee2017desire,
  title={{DESIRE}: Distant future prediction in dynamic scenes with interacting agents},
  author={Lee, Namhoon and Choi, Wongun and Vernaza, Paul and Choy, Christopher B and Torr, Philip HS and Chandraker, Manmohan},
  booktitle={Proceedings of the IEEE Conference on Computer Vision and Pattern Recognition},
  pages={336--345},
  year={2017}
}

@inproceedings{park2023leveraging,
  title={Leveraging future relationship reasoning for vehicle trajectory prediction},
  author={Park, Daehee and Ryu, Hobin and Yang, Yunseo and Cho, Jegyeong and Kim, Jiwon and Yoon, Kuk-Jin},
  booktitle={International Conference on Learning Representations},
  year={2023}
}

@inproceedings{jeong2024multi,
  title={Multi-agent long-term {3D} human pose forecasting via interaction-aware trajectory conditioning},
  author={Jeong, Jaewoo and Park, Daehee and Yoon, Kuk-Jin},
  booktitle={Proceedings of the IEEE/CVF Conference on Computer Vision and Pattern Recognition},
  pages={1617--1628},
  year={2024}
}

@inproceedings{zhou2022hivt,
  title={{HiVT}: Hierarchical vector transformer for multi-agent motion prediction},
  author={Zhou, Zikang and Ye, Luyao and Wang, Jianping and Wu, Kui and Lu, Kejie},
  booktitle={Proceedings of the IEEE/CVF Conference on Computer Vision and Pattern Recognition},
  pages={8823--8833},
  year={2022}
}

@inproceedings{weng2023joint,
  title={Joint metrics matter: A better standard for trajectory forecasting},
  author={Weng, Erica and Hoshino, Hana and Ramanan, Deva and Kitani, Kris},
  booktitle={Proceedings of the IEEE/CVF International Conference on Computer Vision},
  pages={20315--20326},
  year={2023}
}

@inproceedings{scott2022soccertrack,
  title={{SoccerTrack}: A dataset and tracking algorithm for soccer with fish-eye and drone videos},
  author={Scott, Atom and Uchida, Ikuma and Onishi, Masaki and Kameda, Yoshinari and Fukui, Kazuhiro and Fujii, Keisuke},
  booktitle={Proceedings of the IEEE/CVF Conference on Computer Vision and Pattern Recognition Workshops},
  pages={3569--3579},
  year={2022}
}

@inproceedings{zhan2018generating,
  title={Generating multi-agent trajectories using programmatic weak supervision},
  author={Zhan, Eric and Zheng, Stephan and Yue, Yisong and Sha, Long and Lucey, Patrick},
  booktitle={International Conference on Learning Representations},
  year={2019}
}

@misc{nfl_big_data_bowl,
  title        = {{Big Data Bowl}},
  author       = {{National Football League}},
  year         = {2019},
  howpublished = {\url{https://github.com/nfl-football-ops/Big-Data-Bowl}},
  note         = {Accessed: 2025-11-06}
}

@inproceedings{gu2022stochastic,
  title={Stochastic Trajectory Prediction via Motion Indeterminacy Diffusion},
  author={Gu, Tianpei and Chen, Guangyi and Li, Junlong and Lin, Chunze and Rao, Yongming and Zhou, Jie and Lu, Jiwen},
  booktitle={Proceedings of the IEEE/CVF Conference on Computer Vision and Pattern Recognition},
  pages={17113--17122},
  year={2022}
}

@inproceedings{jiang2023motiondiffuser,
  title={{MotionDiffuser}: Controllable multi-agent motion prediction using diffusion},
  author={Jiang, Chiyu and Cornman, Andre and Park, Cheolho and Sapp, Benjamin and Zhou, Yin and Anguelov, Dragomir and others},
  booktitle={Proceedings of the IEEE/CVF Conference on Computer Vision and Pattern Recognition},
  pages={9644--9653},
  year={2023}
}

@inproceedings{rowe2023fjmp,
  title={{FJMP}: Factorized joint multi-agent motion prediction over learned directed acyclic interaction graphs},
  author={Rowe, Luke and Ethier, Martin and Dykhne, Eli-Henry and Czarnecki, Krzysztof},
  booktitle={Proceedings of the IEEE/CVF Conference on Computer Vision and Pattern Recognition},
  pages={13745--13755},
  year={2023}
}

@article{sun2023conformal,
  title={Conformal prediction for uncertainty-aware planning with diffusion dynamics model},
  author={Sun, Jiankai and Jiang, Yiqi and Qiu, Jianing and Nobel, Parth and Kochenderfer, Mykel J and Schwager, Mac},
  journal={Advances in Neural Information Processing Systems},
  volume={36},
  pages={80324--80337},
  year={2023}
}

@inproceedings{neumeier2024reliable,
  title={Reliable trajectory prediction and uncertainty quantification with conditioned diffusion models},
  author={Neumeier, Marion and Dorn, Sebastian and Botsch, Michael and Utschick, Wolfgang},
  booktitle={Proceedings of the IEEE/CVF Conference on Computer Vision and Pattern Recognition Workshops},
  pages={3461--3470},
  year={2024}
}

@inproceedings{capellera2025unified,
  title={Unified uncertainty-aware diffusion for multi-agent trajectory modeling},
  author={Capellera, Guillem and Rubio, Antonio and Ferraz, Luis and Agudo, Antonio},
  booktitle={Proceedings of the Computer Vision and Pattern Recognition Conference},
  pages={22476--22486},
  year={2025}
}

@inproceedings{kim2024higher,
  title={Higher-order relational reasoning for pedestrian trajectory prediction},
  author={Kim, Sungjune and Chi, Hyung-gun and Lim, Hyerin and Ramani, Karthik and Kim, Jinkyu and Kim, Sangpil},
  booktitle={Proceedings of the IEEE/CVF Conference on Computer Vision and Pattern Recognition},
  pages={15251--15260},
  year={2024}
}

@inproceedings{bae2022non,
  title={Non-probability sampling network for stochastic human trajectory prediction},
  author={Bae, Inhwan and Park, Jin-Hwi and Jeon, Hae-Gon},
  booktitle={Proceedings of the IEEE/CVF Conference on Computer Vision and Pattern Recognition},
  pages={6477--6487},
  year={2022}
}

@inproceedings{dendorfer2021mg,
  title={{MG-GAN}: A multi-generator model preventing out-of-distribution samples in pedestrian trajectory prediction},
  author={Dendorfer, Patrick and Elflein, Sven and Leal-Taix{\'e}, Laura},
  booktitle={Proceedings of the IEEE/CVF International Conference on Computer Vision},
  pages={13158--13167},
  year={2021}
}

@inproceedings{zhao2019multi,
  title={Multi-agent tensor fusion for contextual trajectory prediction},
  author={Zhao, Tianyang and Xu, Yifei and Monfort, Mathew and Choi, Wongun and Baker, Chris and Zhao, Yibiao and Wang, Yizhou and Wu, Ying Nian},
  booktitle={Proceedings of the IEEE/CVF Conference on Computer Vision and Pattern Recognition},
  pages={12126--12134},
  year={2019}
}

@inproceedings{huang2023gameformer,
  title={{GameFormer}: Game-theoretic modeling and learning of transformer-based interactive prediction and planning for autonomous driving},
  author={Huang, Zhiyu and Liu, Haochen and Lv, Chen},
  booktitle={Proceedings of the IEEE/CVF International Conference on Computer Vision},
  pages={3903--3913},
  year={2023}
}

@InProceedings{xu2023eqmotion,
  author    = {Xu, Chenxin and Tan, Robby T. and Tan, Yuhong and Chen, Siheng and Wang, Yu Guang and Wang, Xinchao and Wang, Yanfeng},
  title     = {{EqMotion}: Equivariant Multi-Agent Motion Prediction with Invariant Interaction Reasoning},
  booktitle = {Proceedings of the IEEE/CVF Conference on Computer Vision and Pattern Recognition},
  year      = {2023},
  pages     = {1410--1420},
}

@inproceedings{bae2024singulartrajectory,
  title={{SingularTrajectory}: Universal trajectory predictor using diffusion model},
  author={Bae, Inhwan and Park, Young-Jae and Jeon, Hae-Gon},
  booktitle={Proceedings of the IEEE/CVF Conference on Computer Vision and Pattern Recognition},
  pages={17890--17901},
  year={2024}
}

@InProceedings{yang2025unimtd,
  author    = {Yang, Songru and Shi, Zhenwei and Zou, Zhengxia},
  title     = {Unified Multi-Agent Trajectory Modeling with Masked Trajectory Diffusion},
  booktitle = {Proceedings of the IEEE/CVF International Conference on Computer Vision},
  pages     = {27563--27574},
  year      = {2025},
}

@inproceedings{liu2025intdiff,
  author  = {Liu, Yu and Liu, Zhijie and Ren, Xiao and Li, You-Fu and Kong, He},
  title   = {Intention Enhanced Diffusion Model for Multimodal Pedestrian Trajectory Prediction},
  booktitle = {IEEE International Conference on Intelligent Transportation Systems (ITSC)},
  pages   = {2063--2068},
  year    = {2025},
}

@InProceedings{socialmoif2025,
  author    = {Chen, Kai and Zhao, Xiaodong and Huang, Yujie and Fang, Guoyu and Song, Xiao and Wang, Ruiping and Wang, Ziyuan},
  title     = {{SocialMOIF}: Multi-Order Intention Fusion for Pedestrian Trajectory Prediction},
  booktitle = {Proceedings of the IEEE/CVF Conference on Computer Vision and Pattern Recognition},
  year      = {2025},
  pages     = {22465--22475},
}

@inproceedings{robicquet2016learning,
  title={Learning social etiquette: Human trajectory understanding in crowded scenes},
  author={Robicquet, Alexandre and Sadeghian, Amir and Alahi, Alexandre and Savarese, Silvio},
  booktitle={European Conference on Computer Vision},
  pages={549--565},
  year={2016},
  organization={Springer}
}

@article{wang2026demystifing,
  title={Demystifying Video Reasoning},
  author={Wang, Ruisi and Cai, Zhongang and Pu, Fanyi and Xu, Junxiang and Yin, Wanqi and Wang, Maijunxian and Ji, Ran and Gu, Chenyang and Li, Bo and Huang, Ziqi and others},
  journal={arXiv preprint arXiv:2603.16870},
  year={2026}
}

@inproceedings{fu2025moflowonestepflowmatching,
  title={{MoFlow}: One-step flow matching for human trajectory forecasting via implicit maximum likelihood estimation based distillation},
  author={Fu, Yuxiang and Yan, Qi and Wang, Lele and Li, Ke and Liao, Renjie},
  booktitle={Proceedings of the Computer Vision and Pattern Recognition Conference},
  pages={17282--17293},
  year={2025}
}

@article{shi2022motion,
  title={Motion transformer with global intention localization and local movement refinement},
  author={Shi, Shaoshuai and Jiang, Li and Dai, Dengxin and Schiele, Bernt},
  journal={Advances in Neural Information Processing Systems},
  volume={35},
  pages={6531--6543},
  year={2022}
}

@article{shi2024mtr++,
  title={{MTR++}: Multi-agent motion prediction with symmetric scene modeling and guided intention querying},
  author={Shi, Shaoshuai and Jiang, Li and Dai, Dengxin and Schiele, Bernt},
  journal={IEEE Transactions on Pattern Analysis and Machine Intelligence},
  volume={46},
  number={5},
  pages={3955--3971},
  year={2024},
  publisher={IEEE}
}

@inproceedings{jia2023towards,
  title={Towards capturing the temporal dynamics for trajectory prediction: a coarse-to-fine approach},
  author={Jia, Xiaosong and Chen, Li and Wu, Penghao and Zeng, Jia and Yan, Junchi and Li, Hongyang and Qiao, Yu},
  booktitle={Conference on Robot Learning},
  pages={910--920},
  year={2023},
  organization={PMLR}
}

@inproceedings{luo2023jfp,
  title={{JFP}: Joint future prediction with interactive multi-agent modeling for autonomous driving},
  author={Luo, Wenjie and Park, Cheol and Cornman, Andre and Sapp, Benjamin and Anguelov, Dragomir},
  booktitle={Conference on Robot Learning},
  pages={1457--1467},
  year={2023},
  organization={PMLR}
}

@inproceedings{zheng2023layoutdiffusion,
  title={{LayoutDiffusion}: Controllable diffusion model for layout-to-image generation},
  author={Zheng, Guangcong and Zhou, Xianpan and Li, Xuewei and Qi, Zhongang and Shan, Ying and Li, Xi},
  booktitle={Proceedings of the IEEE/CVF Conference on Computer Vision and Pattern Recognition},
  pages={22490--22499},
  year={2023}
}

@inproceedings{farshad2023scenegenie,
  title={{SceneGenie}: Scene graph guided diffusion models for image synthesis},
  author={Farshad, Azade and Yeganeh, Yousef and Chi, Yu and Shen, Chengzhi and Ommer, Bj{\"o}rn and Navab, Nassir},
  booktitle={Proceedings of the IEEE/CVF International Conference on Computer Vision Workshops},
  pages={88--98},
  year={2023}
}

@inproceedings{liu2024r3cd,
  title={{R3CD}: Scene graph to image generation with relation-aware compositional contrastive control diffusion},
  author={Liu, Jinxiu and Liu, Qi},
  booktitle={Proceedings of the AAAI Conference on Artificial Intelligence},
  volume={38},
  pages={3657--3665},
  year={2024}
}

@article{zhai2023commonscenes,
  title={{CommonScenes}: Generating commonsense {3D} indoor scenes with scene graph diffusion},
  author={Zhai, Guangyao and {\"O}rnek, Evin P{\i}nar and Wu, Shun-Cheng and Di, Yan and Tombari, Federico and Navab, Nassir and Busam, Benjamin},
  journal={Advances in Neural Information Processing Systems},
  volume={36},
  pages={30026--30038},
  year={2023}
}

@inproceedings{chen2025diffvsgg,
  title={{DIFFVSGG}: Diffusion-driven online video scene graph generation},
  author={Chen, Mu and Li, Liulei and Wang, Wenguan and Yang, Yi},
  booktitle={Proceedings of the Computer Vision and Pattern Recognition Conference},
  pages={29161--29172},
  year={2025}
}

@inproceedings{hu2025spade,
  title={{SPADE}: Spatial-aware denoising network for open-vocabulary panoptic scene graph generation with long-and local-range context reasoning},
  author={Hu, Xin and Qin, Ke and Duan, Guiduo and Li, Ming and Li, Yuan-Fang and He, Tao},
  booktitle={Proceedings of the IEEE/CVF International Conference on Computer Vision},
  pages={15562--15572},
  year={2025}
}

@inproceedings{capellera2025jointdiff,
  title={{JointDiff}: Bridging continuous and discrete in multi-agent trajectory generation},
  author={Capellera, Guillem and Ferraz, Luis and Rubio, Antonio and Alahi, Alexandre and Agudo, Antonio},
  booktitle={International Conference on Learning Representations},
  year={2026}
}

@inproceedings{pei2025foresight,
  title={Foresight in motion: Reinforcing trajectory prediction with reward heuristics},
  author={Pei, Muleilan and Shi, Shaoshuai and Chen, Xuesong and Liu, Xu and Shen, Shaojie},
  booktitle={Proceedings of the IEEE/CVF International Conference on Computer Vision},
  pages={28303--28312},
  year={2025}
}

@inproceedings{xiao2025srefiner,
  title={{SRefiner}: Soft-Braid Attention for Multi-Agent Trajectory Refinement},
  author={Xiao, Liwen and Pan, Zhiyu and Wang, Zhicheng and Cao, Zhiguo and Li, Wei},
  booktitle={Proceedings of the IEEE/CVF International Conference on Computer Vision},
  pages={960--969},
  year={2025}
}

@inproceedings{tang2024hpnet,
  title={{HPNet}: Dynamic trajectory forecasting with historical prediction attention},
  author={Tang, Xiaolong and Kan, Meina and Shan, Shiguang and Ji, Zhilong and Bai, Jinfeng and Chen, Xilin},
  booktitle={Proceedings of the IEEE/CVF Conference on Computer Vision and Pattern Recognition},
  pages={15261--15270},
  year={2024}
}

@inproceedings{thakkar2025poly,
  title={Poly-Autoregressive Prediction for Modeling Interactions},
  author={Thakkar, Neerja and Sadjadpour, Tara and Rajasegeran, Jathushan and Ginosar, Shiry and Malik, Jitendra},
  booktitle={Proceedings of the Computer Vision and Pattern Recognition Conference},
  pages={12402--12412},
  year={2025}
}

@inproceedings{wen2024density,
  title={Density-adaptive model based on motif matrix for multi-agent trajectory prediction},
  author={Wen, Di and Xu, Haoran and He, Zhaocheng and Wu, Zhe and Tan, Guang and Peng, Peixi},
  booktitle={Proceedings of the IEEE/CVF Conference on Computer Vision and Pattern Recognition},
  pages={14822--14832},
  year={2024}
}

@inproceedings{
    fang2025neuralized,
    title={Neuralized {Markov} Random Field for Interaction-Aware Stochastic Human Trajectory Prediction},
    author={Zilin Fang and David Hsu and Gim Hee Lee},
    booktitle={The Thirteenth International Conference on Learning Representations},
    year={2025},
    url={https://openreview.net/forum?id=r3cEOVj7Ze}
}

@inproceedings{park2024t4p,
  title={{T4P}: Test-time training of trajectory prediction via masked autoencoder and actor-specific token memory},
  author={Park, Daehee and Jeong, Jaeseok and Yoon, Sung-Hoon and Jeong, Jaewoo and Yoon, Kuk-Jin},
  booktitle={Proceedings of the IEEE/CVF Conference on Computer Vision and Pattern Recognition},
  pages={15065--15076},
  year={2024}
}

@inproceedings{park2024improving,
  title={Improving transferability for cross-domain trajectory prediction via neural stochastic differential equation},
  author={Park, Daehee and Jeong, Jaewoo and Yoon, Kuk-Jin},
  booktitle={Proceedings of the AAAI Conference on Artificial Intelligence},
  volume={38},
  pages={10145--10154},
  year={2024}
}

@inproceedings{lee2025interaction,
  title={Interaction-merged motion planning: Effectively leveraging diverse motion datasets for robust planning},
  author={Lee, Giwon and Jeong, Wooseong and Park, Daehee and Jeong, Jaewoo and Yoon, Kuk-Jin},
  booktitle={Proceedings of the IEEE/CVF International Conference on Computer Vision},
  pages={28610--28621},
  year={2025}
}

@inproceedings{lee2025nondifferentiable,
  title={Non-differentiable reward optimization for diffusion-based autonomous motion planning},
  author={Lee, Giwon and Park, Daehee and Jeong, Jaewoo and Yoon, Kuk-Jin},
  booktitle={IEEE/RSJ International Conference on Intelligent Robots and Systems (IROS)},
  pages={13215--13222},
  year={2025}
}
